\documentclass[]{IEEEtran}

\IEEEoverridecommandlockouts                              
\usepackage{amssymb,amsmath}
\usepackage{graphicx}
\usepackage{color}
\usepackage{url}
\usepackage[linesnumbered,lined]{algorithm2e}
\usepackage{subfigure}
\usepackage{wrapfig}
\usepackage{todonotes}
\usepackage{amsthm}
\usepackage{multirow}

\newtheorem{lemma}{Lemma}

\newtheorem{theorem}{Theorem}

\title{\LARGE\bf Combining Geometric and Information-Theoretic Approaches\\ for Multi-Robot Exploration}%
\author{Aravind Preshant Premkumar, Kevin Yu, and Pratap Tokekar%
\thanks{The authors are with the Department of Electrical \& Computer Engineering, Virginia Tech, Blacksburg VA, USA. \texttt{\small \{aravindp, klyu, tokekar\}@vt.edu}}
\thanks{This material is based upon work supported by the National Science Foundation under Grant No. 1566247.}
\thanks{This manuscript extends the work presented at~\cite{preshant2016geometric}. The information-theoretic approach presented in Section~\ref{sec:orient} did not appear in~\cite{preshant2016geometric}.}}

\begin{document}

\thispagestyle{empty}
\pagestyle{empty}

\maketitle

\begin{abstract}
We present an algorithm to explore an orthogonal polygon using a team of $p$ robots. This algorithm combines ideas from information-theoretic exploration algorithms and computational geometry based exploration algorithms. We show that the exploration time of our algorithm is competitive (as a function of $p$) with respect to the offline optimal exploration algorithm. The algorithm is based on a single-robot polygon exploration algorithm, a tree exploration algorithm for higher level planning and a submodular orienteering algorithm for lower level planning. We discuss how this strategy can be adapted to real-world settings to deal with noisy sensors. In addition to theoretical analysis, we investigate the performance of our algorithm through simulations for multiple robots and experiments with a single robot.
\end{abstract}

\section{Introduction}
Exploration of unknown environments using a single robot has been a well-studied problem~\cite{rao1993robot,julia2012comparison}. The task can be performed faster if multiple robots are used. The challenge is to come up with an algorithm to efficiently coordinate multiple robots to explore the environment in the minimum amount of time. There have been two types of approaches towards solving the exploration problem: geometric and information-theoretic. In geometric approaches (e.g.,~\cite{bhattacharya2001exploring}), it is typical to assume that the robots have perfect sensing. Geometric algorithms typically give global guarantees on distance traveled at the expense of restrictive assumptions about the environment and sensor models. On the other hand, information-theoretic approaches (e.g.,~\cite{charrow2015information}) explicitly take into account practical constraints such as noisy sensors and complex environments. However, these approaches are often greedy (e.g., frontier-based~\cite{yamauchi1997frontier}) and typically do not yield any guarantees on the total time taken. In this paper, we investigate the challenges in combining information-theoretic algorithms with geometric exploration algorithms while preserving guarantees on exploration time.

We use competitive analysis~\cite{motwani2010randomized} in order to analyze the cost of exploration. Competitive ratio of an online algorithm is defined as the worst-case ratio (over all inputs) of the cost of the online algorithm and the optimal offline algorithm cost. The optimal offline algorithm corresponds to the case when the input (i.e., the map of the environment) itself is known. The goal is to find online algorithms with small, constant competitive ratios, i.e., algorithms whose online performance is comparable to algorithms who know the input a priori.

We focus on the case of exploring unknown orthogonal polygons\footnote{An orthogonal polygon is one in which all sides are aligned with either the $X$ or the $Y$ axes.} without any holes. Deng et al.~\cite{deng1998learn} showed that there is an algorithm with a constant competitive ratio for exploring orthogonal 2-dimensional polygons with a single robot. We present an algorithm with constant competitive ratio for exploring with $p$ robots, when $p$ is fixed (Section~\ref{sec:algo}).

The analysis of this algorithm requires certain assumptions that do not necessarily hold in practice. Our second contribution is to show how to adapt this purely geometric algorithm for real-world constraints to incorporate sensing limitations and uncertainty (Section~\ref{sec:adapt}). We then extend this algorithm in order to improve the quality of the resulting map. We add a local planner to our algorithm that optimizes the information gain of the path taken by the robots while traversing in order to reduce the overall entropy in the map. We evaluate our algorithm through simulations and experiments on a mobile robot (Sections~\ref{sec:sims} and \ref{sec:expt}). We begin with a discussion of the related work.

\section{Related Work}
In this section, we present the existing work related to the exploration problem. We organize the related work into three broad categories: polygon exploration, graph exploration, and information-theoretic exploration.

\subsection{Polygon Exploration}
The study of geometric problems that are based on visibility is a well-established field within computational geometry. The classic problems are the art gallery problem~\cite{o1987art}, watchman route problem~\cite{chin1988optimum}, target search~\cite{fleischer2008competitive} and shortest path planning~\cite{papadimitriou1991shortest} in unknown environments.

Using a fixed set of positions for guarding a known $n$-sided polygonal region, i.e., a set of points from which the entire polygon is visible, is known as the classical art gallery problem. Chvatal~\cite{chvatal1975combinatorial} and Fisk~\cite{fisk1978short} proved that $\lfloor n/3 \rfloor$  guards are always sufficient and sometimes necessary to cover a polygon of $n$ sides. The minimum number of guards required for a specific polygon may be much smaller than this upper bound. 
However, Schuchardt and Hecker \cite{schuchardt1995two} showed that finding a minimum set of guards is NP-hard, even for the special case of an orthogonal polygon.
    
Finding the shortest tour along which one mobile guard can see the polygon completely is the watchman route problem. Chin and Ntafos \cite{chin1988optimum} showed that the watchman route can be found in polynomial time in a simple orthogonal polygon. Wang et al.~\cite{wang2010generalized} showed that the watchman route problem is for general environments is NP-hard and presented a $\mathcal{O}(\text{polylog} n)$ approximation for the restricted case when each viewpoint is required to see a complete polygon side. 

Exploring an unknown polygon is the online watchman route problem. Bhattacharya et al.~\cite{bhattacharya2001exploring} and Ghosh et al.~\cite{ghosh2008online} approached the exploration problem with discrete vision, i.e., they assume that the robot does not have continuous visibility and has to stop at different scan points in order to sense the environment. They focus on the worst-case number of necessary scan points. Their algorithm results in a competitive ratio of $(r +1)/2$, where $r$ is the number of reflex vertices in the polygon. For limited range of visibility, they give an algorithm where the competitive ratio in a polygon $P$ can be limited by $\lfloor \frac{8\pi}{3}+ \frac{\pi R \times Perimeter(P)}{Area(P)}  + \frac{(r+h+1)\pi R^{2}}{Area(P)}\rfloor$, where $h$ is the number of holes and $R$ is the number of reflex vertices in the polygon.

Albers et al.~\cite{albers2000exploring} assume that the environment is modeled by a directed, strongly connected graph and give a $d^{\mathcal{O} \log d}m$ competitive algorithm where $m$ is the number edges in the graph and $d$ is the minimum number of edges that have
to be added to make the graph Eulerian. The robot's task is to visit all nodes and edges of the graph using the minimum number $R$ of edge traversals. For a simple polygon, Hoffmann et al.~\cite{hoffmann2001polygon} presented an algorithm which achieves a competitive ratio of $26.5$. For the special case of an orthogonal polygon, Deng et al.~\cite{deng1998learn} presented a $\sqrt{2}$ competitive exploration strategy with one robot. We show how to extend the single robot exploration algorithm by Deng et al.~\cite{deng1998learn} to the case of $p$ robots. The resulting algorithm has a competitive ratio that is a function of $p$.
    
\subsection{Graph Exploration}
The problem of visiting all the nodes in a graph in the least amount of time is known as the Traveling Salesperson Problem (TSP). Here, all nodes of the graph are known before-hand and the objective is to determine the shortest tour visiting all the nodes in the graph exactly once. Finding the optimal TSP tour for a given graph is known to be NP-hard, even for the special case where the nodes in the graph represent points on the Euclidean plane~\cite{mitchell1999guillotine}. For the Euclidean version of the problem, there exist polynomial time approximation schemes~\cite{mitchell1999guillotine,arora1998polynomial}, i.e., for any $\epsilon > 0$, there exists a
polynomial time algorithm which guarantees an approximation factor of $1 + \epsilon$. 

In the graph exploration version of the problem, nodes are revealed in an online fashion. The objective is to minimize the total distance (or time) traveled. Fraigniaud et al.~\cite{fraigniaud2006collective} presented an algorithm for exploration of trees using $p$ robots with a competitive ratio of $\mathcal{O}(p/ \log p)$ and a lower bound of $\Omega(2 + 1/p)$. This lower bound was improved by Dynia et al.~\cite{dynia2007robots} to $\Omega(\log p / \log\log p)$. They modeled the cost as the maximal number of edges traversed by a robot and presented a $(4-2/p)$ competitive online algorithm. Higashikawa et al.~\cite{higashikawa2014online} showed that greedy exploration strategies have an even stronger lower bound of $\Omega(p/ \log p)$ and presented a $(p + \log p/1+\log p)$ competitive algorithm.

Better bounds have been achieved for restricted graphs. Dynia et al.~\cite{dynia2006smart} presented an algorithm that achieves faster exploration for trees restricted by a density parameter $k$ which forces a minimum depth for any subtree depending on its size. Trees embeddable in $d$-dimensional grids can be explored with a competitive ratio of $\mathcal{O}(d^{1-1/k})$. For 2-dimensional grids with only convex obstacles, Ortolf et al.~\cite{ortolf2012online} improved the competitive ratio to $\mathcal{O}(\log^{2} d)$. Despite these strong restrictions on the graph, the same lower bound of $\Omega( \log p/\log \log p )$ holds for all trees.

We show that the problem of exploring a polygon can be formulated as a multi-robot tree exploration problem. Our algorithm yields a competitive ratio of $\frac{2(\sqrt2{p} +{\log p})}{1+\log p}$ where $p$ is the number of robots.

\subsection{Information-Theoretic Exploration}
Geometric and graph-based approaches typically assume perfect sensing with no noise. In practice, however, measurements are not perfect and we have to account for measurement noise when exploring the environment. Such exploration strategies can be broadly classified into frontier-based and information-theoretic strategies~\cite{julia2012comparison}. 

\begin{figure}[htb]
\centering
\includegraphics[scale=0.2, angle=0]{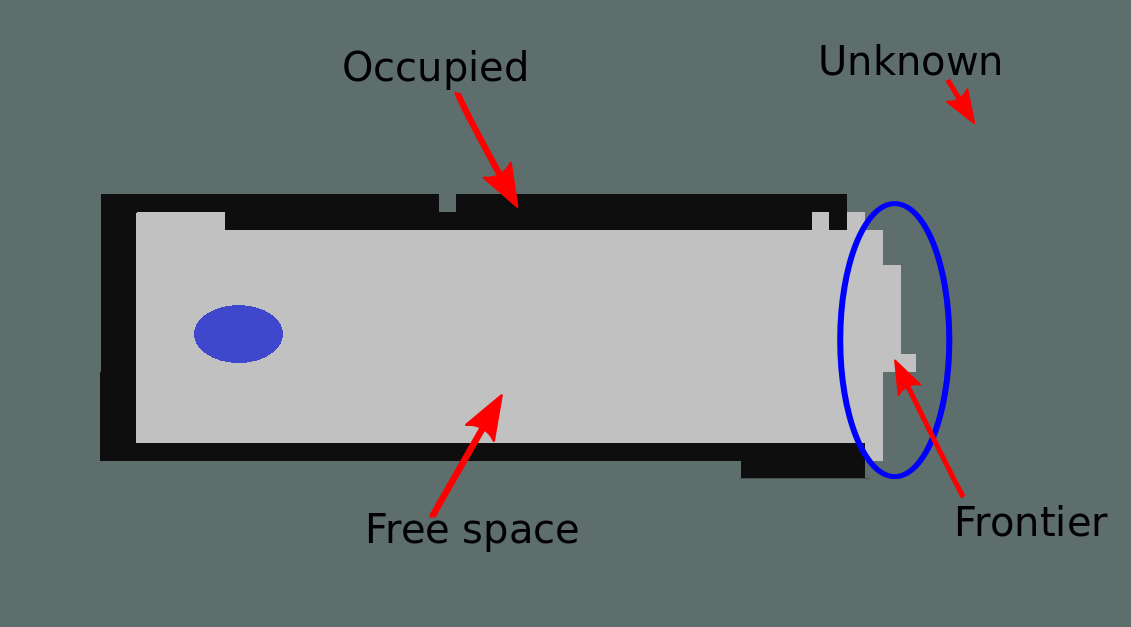}
\caption{Frontier is the boundary between known free space and unknown space.}
\label{fig:frontier}
\end{figure}

Information-Theoretic exploration strategies typically use an occupancy grid to represent the maps generated from noisy and uncertain sensor measurement~\cite{elfes1989using}. An occupancy grid is a uniformly-spaced grid, with a binary random variable (per grid cell) representing the probability of the cell being occupied by an obstacle. There are many existing occupancy grid mapping techniques like gmapping~\cite{stachniss2007gmapping}, octomap~\cite{hornung2013octomap}, RTAB-Map~\cite{labbe2014online} and ORB-slam~\cite{mur2015orb}. Frontier-based exploration strategies are largely greedy in nature and drive the robots to the boundary between known and unknown spaces in the map. In occupancy grids, frontiers are cells determined to be free (probability of occupancy close to zero) which are next to grid cells that have not been observed (probability of occupancy is set to be 0.5), shown in Figure~\ref{fig:frontier}. Variants of this strategy have been used to perform exploration of unknown 2D \cite{yamauchi1997frontier,burgard2005coordinated,schwager2011multi} and 2.5D environments \cite{cesare2015multi}. We refer the reader to the comprehensive study by Holz et al.~\cite{holz2011comparative} for further details.

Information-theoretic strategies seek to optimize some information measure such as entropy (uncertainty)~\cite{whaite1997autonomous,moorehead2001autonomous} in the environment, or mutual information~\cite{amigoni2010information} while exploring the environment. Mutual information~\cite{cover2012elements} between two random variables $X$ and $Y$ is given by, $$ I(X;Y) = \sum_{x} \sum_{y} f(x,y) ~\log \frac{f(x,y)}{f_{1}(x)f_{2}(y)}.$$ Mutual information in occupancy grid predicts how much future measurements will decrease the robot’s uncertainty associated with all grid cells. Julian et al.~\cite{julian2014mutual} studied the computation of mutual information for range based sensors. While these algorithms produce better maps, the planner is typically a one-step greedy algorithm or finite-horizon planners that cannot give global guarantees on the total distance traveled. 

In order to overcome this, a better approach may be to combine a global planner along with a local planner. Davis et al. \cite{davis2016c} proposed one such algorithm where the goal was to maximizes the coverage of a user-specified region while minimizing the control costs of the robot. The authors are given a start state and a goal state and minimize the probability of collision with the environment as well. Bai et al. \cite{bai2016information} used an approach to predict mutual information using Bayesian optimization in which the long-term goal is to reduce entropy throughout the robot's environment map, and the short-term goal is to perform the sensing action in each iteration that will maximize mutual information. Choudhury et al. \cite{choudhury2016learning} showed that supervised learning could be used to predict informative actions without evaluating the expected mutual information exhaustively for every possible action. Charrow et al.~\cite{charrow2015information} attempted to resolve this with a heuristic that uses a global planner for a single robot to determine trajectories that maximizes the Information-Theoretic objective whilst employing a gradient-based trajectory optimization technique to locally refine the chosen trajectory such that the mutual information objective is maximized. We build on this work and present a two-level planner that maximizes information locally while giving strong global performance guarantees on the path length followed by the robots.


\section{Multi-Robot Geometric Exploration Algorithm} \label{sec:algo}

In this section, we present the details of our algorithm for exploring an orthogonal polygon without any holes, $P$, using a team of $p$ robots. Our algorithm builds on the algorithm by Deng et al.~\cite{deng1998learn} for exploring an orthogonal polygon with a single robot and extends it to the case of multiple robots using the graph exploration strategy from \cite{higashikawa2014online}. Our main insight is to show that the path followed by the robot using the algorithm in~\cite{deng1998learn} can be used to construct a tree, denoted by $\mathcal{T}$, in $P$. That is, exploring the polygon is equivalent to visiting all nodes in this tree. We show that a multi-robot tree exploration algorithm from \cite{higashikawa2014online} can be used to explore and visit every node in this tree. Furthermore, we show that the competitive ratio of our algorithm is bounded (as a function of $p$).

We assume all robots start at a common location. The cost of exploration is defined as the time taken for all the robots to return to the starting location having explored the polygon. We say that a polygon $P$ is explored if all points in its interior and on the boundary were seen from at least one robot. For the purpose of the analysis, we assume that the sensor on the robot is an omnidirectional camera with infinite sensing range which returns the exact coordinates of any object in its field of view. We also assume that the robots move at unit speeds and can communicate at all times. In the next section, we show how to adapt our algorithm to realistic sensing models and evaluate it through experiments. 

\begin{figure}[htb]
\centering
\includegraphics[width=0.3\textwidth]{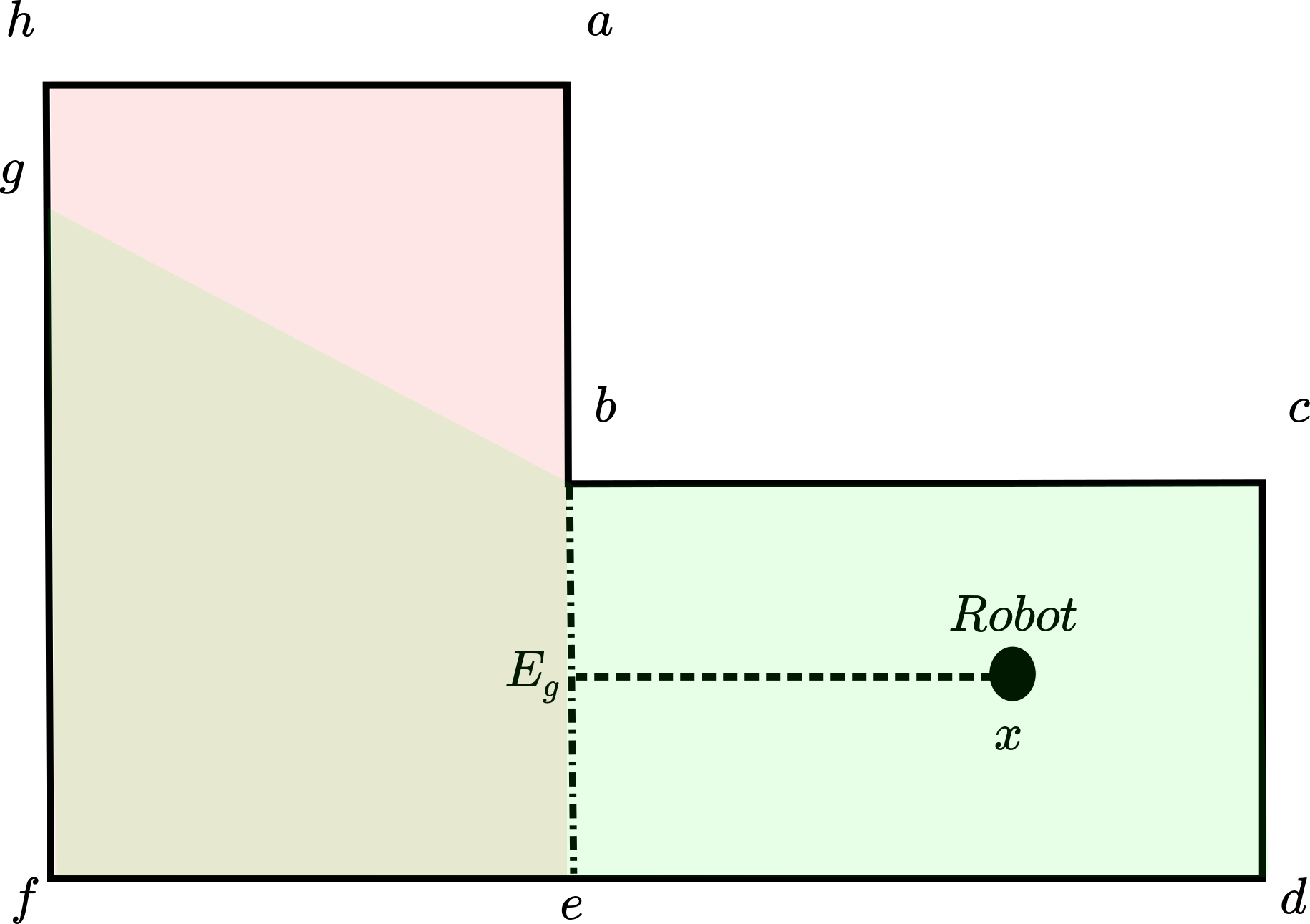}
\caption{Vertex $b$ blocks the visibility of the robot and is known as a \emph{blocking vertex}. The robot has to cross the segment $\overline{be}$ (known as the \emph{extension}) to increase the visible boundary of the polygon $P$. The \emph{extension} divides $P$ into two sub-polygons. The  sub-polygon not containing the robot (overlap of red and green areas) is known as the \emph{foreign polygon} w.r.t. $b$ and is denoted as $FP(b)$.}
\label{visiblity}
\end{figure}

We introduce some terminology used in our algorithm (refer to Figure~\ref{visiblity}) before presenting the details.
The sub-polygon that is visible from a point $x$ is known as the \emph{visibility polygon} of $x$ and is denoted by $VP(x)$. 
Some of the edges in the visibility polygon are part of the boundary of $P$ whereas others are chords in the interior of $P$ (e.g., segment $\overline{gb}$ in Figure~\ref{visiblity}).
A reflex vertex of $P$ which breaks the continuity of the part of the boundary of $P$ visible from $x$ is known as a \emph{blocking vertex}. Vertex $b$ in Figure~\ref{visiblity} is a blocking vertex. 
Let $\overline{bc}$ be the side incident to $b$ which is (partly) visible from $x$. The line segment perpendicular to $\overline{bc}$ drawn from $b$ till the boundary of $P$  is known as the \emph{extension} of the blocking vertex $b$ for an orthogonal polygon. The extension $E_b$ is said to be \emph{associated} with the side $\overline{ab}$
The robot must cross the extension in order to ``look beyond'' the blocking vertex and explore the polygon.

An extension divides $P$ into two sub-polygons. The one which contains the starting position is known as the \emph{home polygon} with respect to $b$. The other sub-polygon is known as the \emph{foreign polygon} with respect to $b$ and is denoted as $FP(b)$. The robot must cross the extension $E_b$ in order to see the side, $\overline{ab}$, that is associated with it, since $\overline{ab}$ lies in the foreign polygon. If the foreign polygon of blocking vertex, $b'$, is completed contained in the foreign polygon of $b$, then $b'$ is said to be dominated by $b$. If a blocking vertex is not dominated by any other blocking vertex, then the corresponding extension is said to be a \emph{critical} extension. Deng et al.~\cite{deng1998learn} showed that a path that visits every critical extension is sufficient to explore the entire polygon.

Draw the line segment starting from the robot's position $x$ perpendicular to the extension $\overline{be}$. The point at which these two line segments intersect ($E_{g}$) is known as the \emph{extension goal} corresponding to the blocking vertex $b$. 

\begin{algorithm}
\caption{Multi-Robot Exploration Subroutine \label{algo}}
\SetKwFunction{explore}{explore}
  \SetKwProg{fn}{Function}{}{}
  \fn{\explore{}}{
  \KwData{Cluster of robots, $C$, located at some node, $A$, in the tree.}
  \eIf{$A$ is marked \emph{under-exploration}}{
      $\mathcal{A}\leftarrow$ children of $A$ not marked as \emph{explored}\; 
      \eIf{$\mathcal{A}==\emptyset$}{
          \eIf{$A==$ root of the tree}{
          Terminate exploration\;
          }{
          \texttt{goto($C$,parent($A$)}\;
          }
      }{
      Divide $C$ equally among $\mathcal{A}$\;
      When any cluster reaches a child of $A$, call \texttt{explore}\;
  }
  }{
    Mark $A$ as \emph{under-exploration}\;
\eIf{blocking vertices detected from $A$}{
    Sort in clockwise direction and add as children of $A$\;
    \For{$a$ and $b$ are distinct blocking vertices}{
        \eIf{$FP(a) \subseteq  FP(b)$}{
            add $a$ as child of $b$\;
        }{ \If{$FP(a)$ and $FP(b)$ intersect}{
            add the vertex that appears first in the clockwise order as the parent of the vertex that appears later;
            }
        }
    }
    Divide $C$ equally among children of $A$\;
    When any cluster reaches a child of $A$, call \texttt{explore}\;
}{
        Mark $A$ as \emph{explored}\;
        \texttt{goto($C$,parent($A$))}\;
    }
}
}
\end{algorithm}

The algorithm starts by creating a tree with the initial position of the robots as the root. All robots start in one cluster located at the root node. We use three labels to keep track of the status of any node in the tree: \emph{unexplored}, \emph{under-exploration}, and \emph{explored}. Whenever a cluster of robots reach a node (using a subroutine \texttt{goto}), we call the subroutine shown in Algorithm~\ref{algo}. The subroutine \texttt{goto} finds the zig-zag-to-point paths between two points, as described by Deng et al.~\cite{deng1998learn}. In addition, if two clusters run into each other, then they merge and travel up the tree together. Furthermore, whenever the \texttt{goto} subroutine is called with a parent node as the destination, the robots do not need to physically travel up to the parent. Instead, they can directly go to the next unexplored node using standard short-cutting techniques.

The root is initially marked as \emph{under-exploration}. We then check to determine any blocking vertices visible from the current node. We add the \emph{extension goals} corresponding to any blocking vertices visible from the current node as its children. All of the corresponding extension goals are added as children by sorting them in the clockwise direction. These new children of the current node are adjusted according to the conditions mentioned. These conditions define an ordering over the nodes, by rewiring the tree. The cluster of robots at the current node is then divided as equally as possible and sent to visit the children of the current node. If the current node doesn't have any children, the current node is marked as \emph{explored}. The cluster moves to the parent of the current node to explore any of its other children which are \emph{unexplored} or \emph{under-exploration}. If a node does not have any children that are \emph{unexplored} or \emph{under-exploration}, then that node and its sub-tree are said to be explored. The exploration is said to be completed if the root of the tree is marked as \emph{explored}.

\begin{figure}[t]
\centering
\includegraphics[scale=0.05, angle=0]
{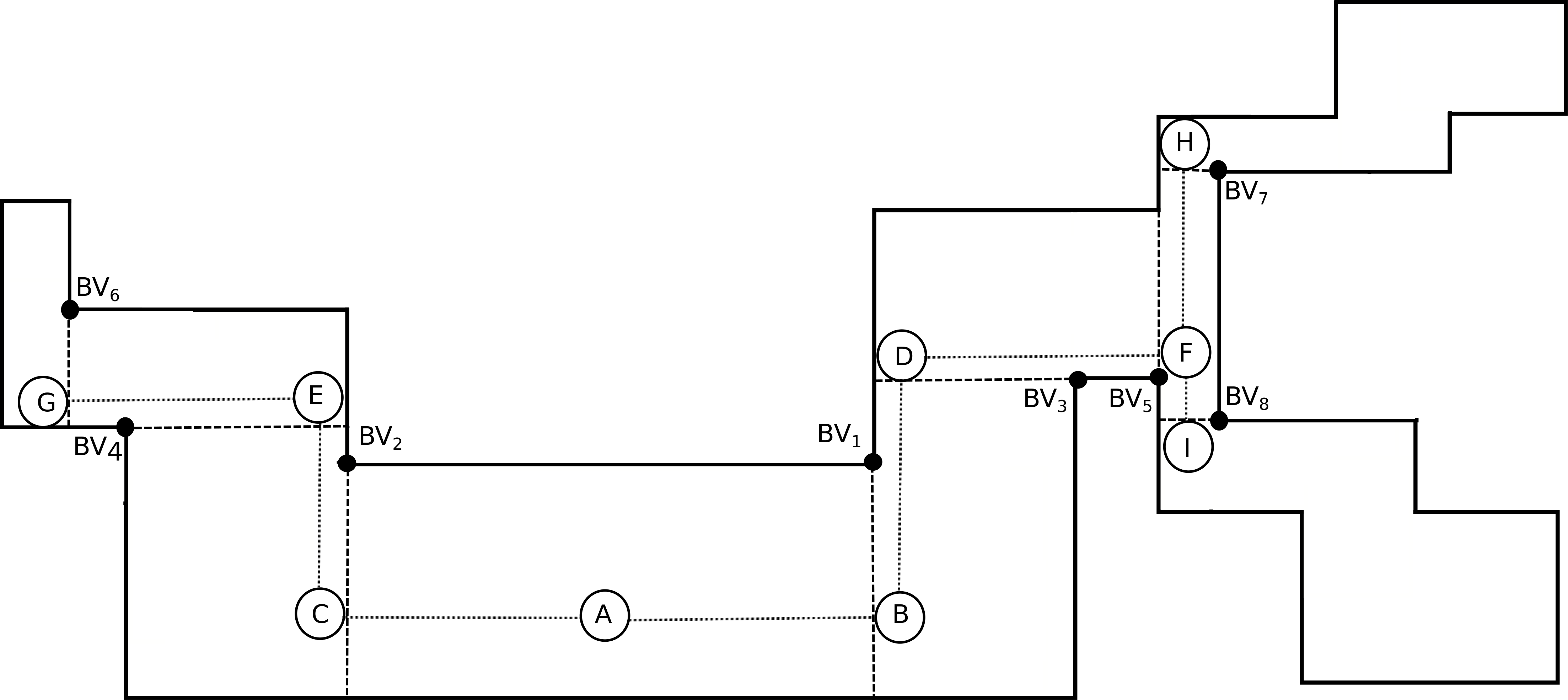}
\caption{An intermediate stage of exploration of a polygon by a team of 4 robots. Two robots located at the location $G$ and the other two are located at $F$.}
\label{intermediate_exploration}
\end{figure}

\begin{figure}[t]
\centering
\includegraphics[scale=0.10, angle=0]
{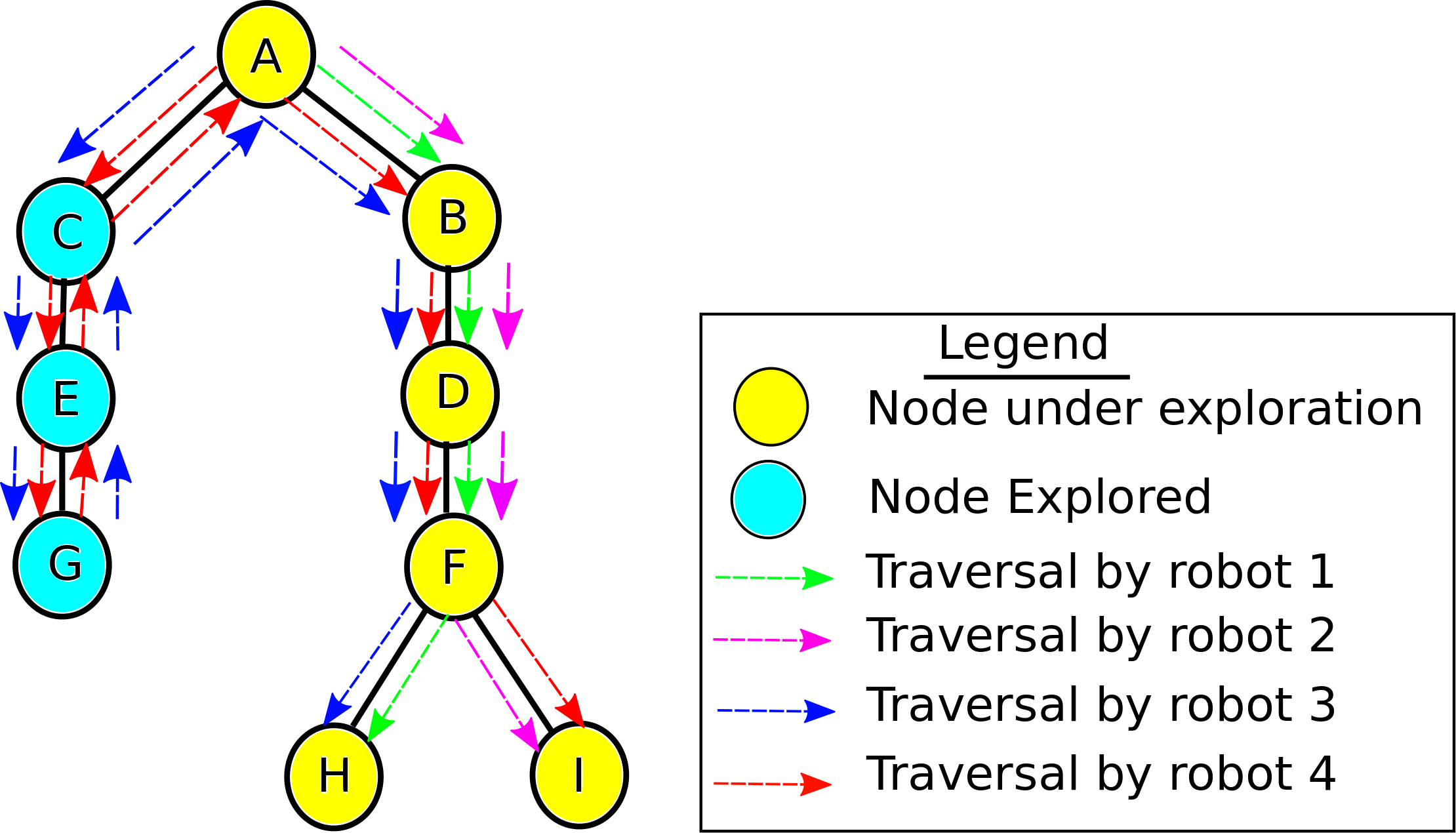}
\caption{The tree corresponding to the stage of exploration shown in Figure~\ref{intermediate_exploration}.}
\label{intermediate_exploration_tree}
\end{figure}

Consider an example where $P$ has been explored partially by a team of four robots as shown in Figure~\ref{intermediate_exploration}. All four robots $\{r_1,r_2,r_3,r_4\}$ start off at the location $A$ which is added as the root of the tree as shown in Figure~\ref{intermediate_exploration_tree}. Node $A$ is marked as \emph{under-exploration}. The robots observe two blocking vertices, $BV_{1}$ and $BV_{2}$. The extension goals $B$ and $C$ corresponding to  $BV_{1}$ and $BV_{2}$, respectively, are added as the children of $A$ in the tree. The robots split into two clusters $\{r_1,r_2\}$ and $\{r_3,r_4\}$. Cluster $\{r_1,r_2\}$ moves towards $B$ and cluster $\{r_3,r_4\}$ moves towards $C$. The corresponding nodes in the tree are marked as \emph{under-exploration}. At $B$, cluster $\{r_1,r_2\}$ observes a new blocking vertex, $BV_{3}$, and the corresponding extension goal $D$ is added as the child of $B$. Since there is only one child, the cluster does not split and both robots move towards $D$. The corresponding node, $D$, in the tree is marked as \emph{under-exploration}. 

At $C$, the cluster $\{r_3,r_4\}$ observes a blocking vertex, $BV_{3}$, and the corresponding extension goal $E$ is added as the child of $C$. Similarly, $F$ is added as the child of $D$ and $G$ is added as the child of $E$. At $F$, cluster $\{r_1,r_2\}$ observes two blocking vertices, the cluster splits into two clusters $\{r_1\}$ and $\{r_2\}$. Cluster $\{r_1\}$ moves towards $H$ and cluster $\{r_2\}$ moves towards $I$. At $G$, cluster $\{r_3,r_4\}$ observes that there are no more blocking vertices. Hence, no children are added to $G$. $G$ is marked as \emph{explored} and the algorithm checks the predecessor in the tree, $E$. Since $E$ does not have any other children for exploration, $E$ is marked as \emph{explored} as well. 

Similarly, the algorithm checks its predecessor, $C$, and marks it as \emph{explored}. Now the algorithm checks $A$, it has a child $B$ which is \emph{under-exploration}. Since there is no blocking vertex, the algorithm proceeds to check $B$. Similarly, the algorithm proceeds to check $D$ and subsequently $F$. At $F$, there are two blocking vertices since neither of the two clusters, $\{r_1\}$ or $\{r_2\}$, have reached their goals. Thus, cluster splits into two $\{r_3\}$ and $\{r_4\}$. $H$ is assigned to $\{r_3\}$ and $I$ is assigned to $\{r_4\}$. The exploration algorithm proceeds in this manner up until the root is marked as \emph{explored}.

\subsection{Competitive Ratio Analysis}
In this section, we prove that the competitive ratio of our algorithm is bounded with respect to the offline optimal algorithm. We divide our analysis into three steps. First, we show that the paths followed by all the robots can be mapped to navigating on a tree. Next, we bound the sum of the costs of edges in the tree with respect to the offline optimal cost. Finally, we bound the cost of our algorithm with respect to the cost of the tree. The graph created by the algorithm is a tree by construction because once a node is added to the graph it is not added to the graph again.

When $p=1$, the proposed algorithm is the same as the one given by Deng et al.~\cite{deng1998learn} for orthogonal polygons without holes. They showed that the competitive ratio of this algorithm is $\sqrt{2}$. The analysis relies on showing that the robot visits the critical extensions in the clockwise order in which their associated sides appear on the boundary of the polygon. We refer the reader to Lemma 6 in~\cite{deng1998learn} for a proof and detailed exposition. 

Let $C_\text{OPT}^p$ denote the time taken by the optimal algorithm for $p$ robots to explore $P$. Let $C_\text{RECT}$ denote the time taken by the strategy from \cite{deng1998learn} for a single robot to explore $P$. We have $C_\text{RECT} \leq \sqrt{2} C_\text{OPT}$ from Theorem 3 in \cite{deng1998learn} and the assumption that the robots travel with unit speeds. Let $C_\text{EXPLORE}^{p}$ be the time taken by the proposed algorithm. We show that the ratio, $C_\text{EXPLORE}^{p}/C_\text{OPT}$, is constant for a given $p$. We will show this by relating both quantities with $C_\text{TREE}^{p}$ which is the sum of the lengths of edges in the tree created by $p$ robots. Note that $C_\text{TREE}^{p}$ is not the same as $C_\text{EXPLORE}^{p}$ since it does not take into account the backtracking that may be involved in the robots' paths.

\begin{lemma}
$C_\text{TREE}^{p} \le \sqrt{2} p C_\text{OPT}^p.$
\end{lemma}
\begin{proof}
When a node is added to the tree (Line 20 in Algorithm~\ref{algo}) it is initially unmarked. When a robot reaches a leaf node, it first marks it as \emph{under-exploration} (Line 15) and then immediately marks it as \emph{explored} (Line 30). Furthermore, all critical extensions will appear as leaf nodes in the tree (due to the rewiring in Lines~18--25). 

Let $T$ be the time taken for the last leaf node to be marked. We have $C_\text{TREE}^{p} \le p T$ since $T$ involves backtracking costs along the tree, whereas $C_\text{TREE}^{p}$ only considers the sum of the edge costs. 

From Lemma 6 in~\cite{deng1998learn}, we know that when $p=1$, the robot visits all the critical extensions in the clockwise order in which their associated sides appear on the boundary of the polygon. From Theorem 2 and Lemma 4 in~\cite{deng1998learn} it follows that if the critical extensions are visited in this clockwise order (and using rectilinear motion in the \texttt{goto} subroutine), then the cost of the tour, $C_\text{EXPLORE}^1$, is no more than $\sqrt{2} C_\text{OPT}^1$. When $p>1$, one robot may not necessarily visit all the critical extensions. However, it still holds that the all critical extensions visited by a robot follow the clockwise order of their associated sides. As a result, we have $T\le \sqrt{2} C_\text{OPT}^p$. Thereby yielding, $C_\text{TREE}^{p} \le \sqrt{2} p C_\text{OPT}^p$.
\end{proof}

\begin{theorem}
\label{Exploration}
If $C_\text{EXPLORE}^{p}$ is the cost of exploring the polygon using the proposed strategy and $C_\text{OPT}^{p}$ is the cost of exploring the polygon using an optimal offline strategy, then we have,
\begin{align} \label{cost_rect}
\frac{C_\text{EXPLORE}^{p}}{C_\text{OPT}^{p}}\le \frac{2(\sqrt2{p} +{\log p})}{1+\log p}.
\end{align}
\end{theorem}
\begin{proof}
The cost of exploring a tree with a recursive depth-first strategy used in the proposed algorithms is given by:
 \begin{align}
 C_\text{EXPLORE}^{p}\le \frac{2(C_\text{TREE}^{p} + d_{\max}{\log p})}{1+\log p},
 \end{align}
 where $d_{\max}$ is the maximum distance of a leaf node from the root in the tree. This bound comes directly from the result in \cite{higashikawa2014online}. In our case, $d_{\max}$ corresponds to the maximum distance of any extension goal from the starting position of the robots. It is easy to see that $d_{\max} \leq C_\text{OPT}^p$.
 
From the previous lemma we have, we have:
\begin{align}
C_\text{EXPLORE}^{p}\le \frac{2(\sqrt2{p}C_\text{OPT}^{p} + C_\text{OPT}^{p}{\log p})}{1+\log p}\nonumber,
\end{align}
which yields
\begin{align} 
 \frac{C_\text{EXPLORE}^{p}}{C_\text{OPT}^{p}}&\le \frac{2(\sqrt2{p} +{\log p})}{1+\log p}.
 \end{align}
\end{proof}

\begin{figure}[htb]
\centering
\includegraphics[width=0.45\textwidth]
{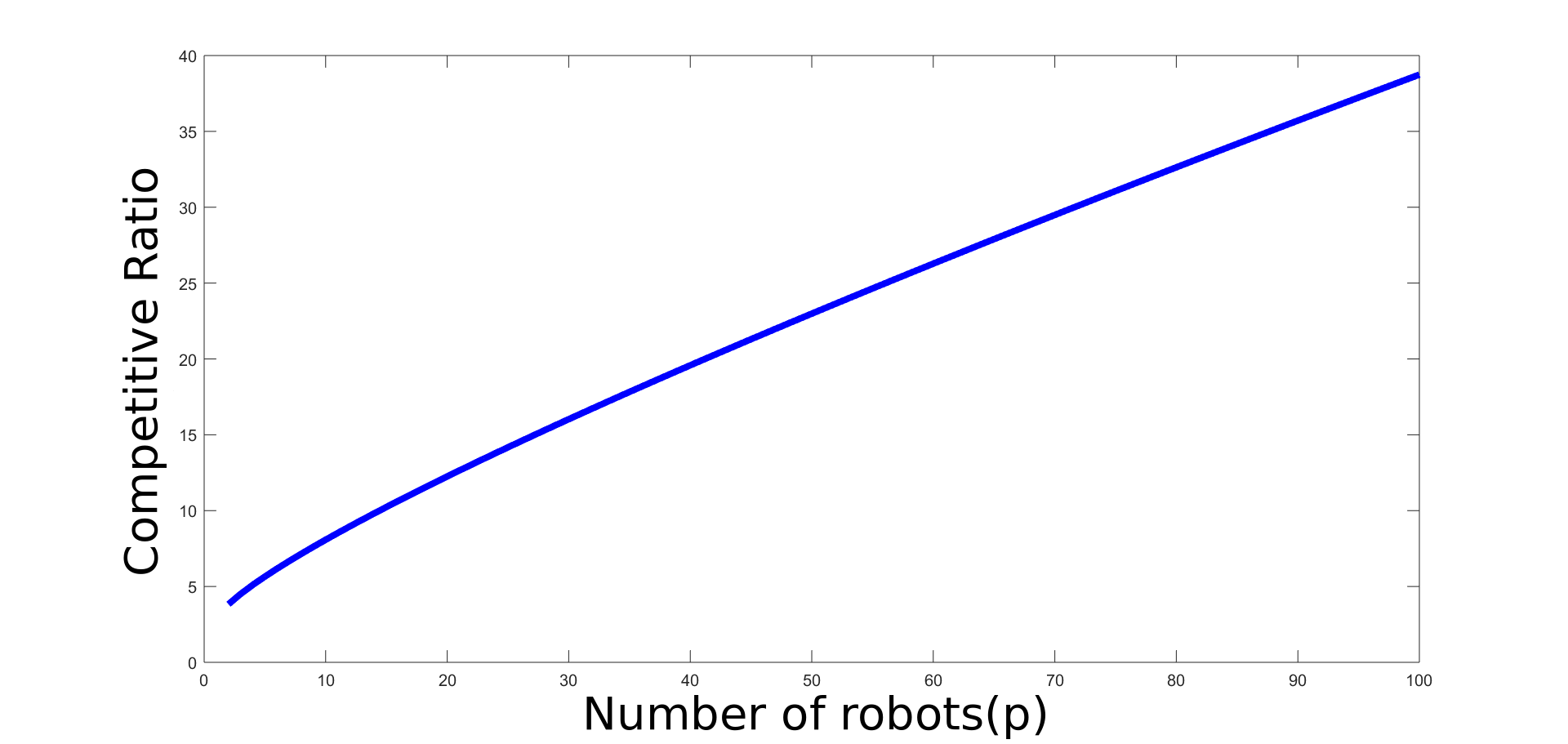}
\caption{Plot of the growth of competitive ratio as a function of the number of robots.\label{competitiveRatio}}
\end{figure}
Thus, we show that the competitive ratio is bounded as a function of $p$. Figure~\ref{competitiveRatio} shows a plot of this bound as a function of $p$. We note that while the analysis only holds for the case of an orthogonal polygon without holes, the resulting algorithm can also be applied for polygons with holes. However, in such a case, the underlying graph created by the robots is not guaranteed to be a tree. Consequently, we would have to use analogous algorithms for exploring general graphs with multiple robots to yield a similar competitive ratio. In the simulations, we show the empirical performance of our algorithm in environments with holes.

\section{Incorporating Information-Theoretic Planning} \label{sec:adapt}

Some of the assumptions made for the analysis do not hold in most practical scenarios. In this section, we show how to extend our basic algorithm framework in order to incorporate real-world constraints.

\subsection{Finding Blocking Vertices in Occupancy Maps}

The main assumption is that of unlimited sensing range. In practice, the robot has a limited sensing range. For example, the robot cannot sense a long corridor with a single observation. Thus, in addition to blocking vertices, the robot has to move to \emph{frontiers} at the end of its sensing range to sense more of the environment. This increases the distance the robot would have to cover compared to the distance it would have to cover if it had an infinite sensor. The robot thus has to detect two types of frontiers, one due to blocking vertices and the other due to sensing range. The only change to the algorithm is in Line 1 where we check for blocking vertices as well as frontiers due to sensing range.

We first detect blocking vertices in a given scan (as described below). Then frontier cells are clustered together to form frontiers due to sensing range. Any frontier which has a constituent frontier cell neighboring a blocking vertex is then discarded. For a blocking vertex, the \emph{extension goal} is added on its \emph{extension} with a slight offset. For frontiers due to sensing range, the middle frontier cell, after clustering, is chosen as the \emph{frontier goal}.

\begin{figure}[htb]
\centering
\includegraphics[height=1.25in]{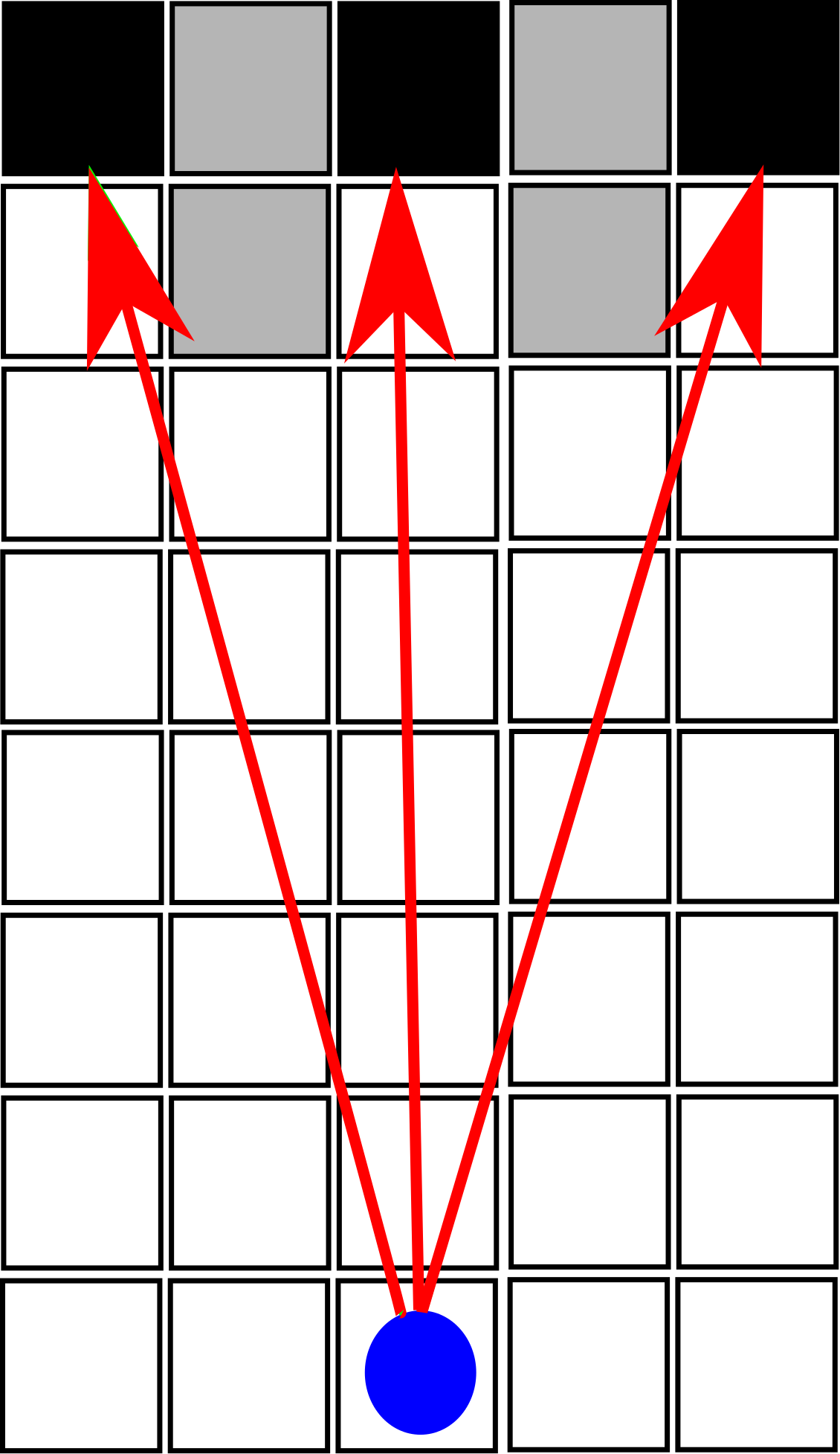}\quad\quad
\includegraphics[height=1.25in]{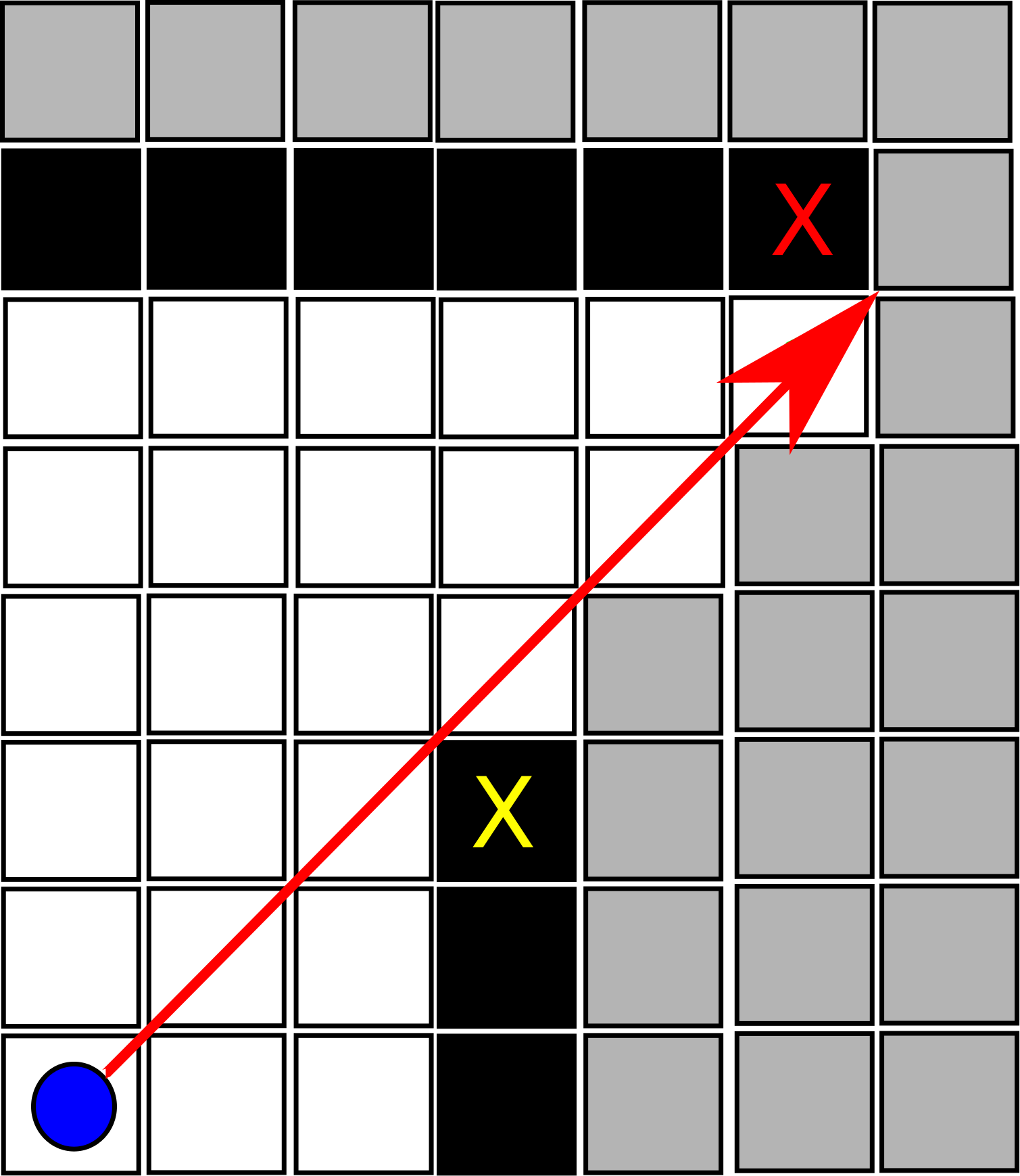}
\caption{\textbf{(Left)} Finite resolution of the laser leads to gaps in observations. \textbf{(Right)} A blocking vertex (marked with yellow `X') has four distinctive neighbors: an occupied cell, an unknown cell, a free cell which is a frontier, one which is not.}
\label{fig:finite_resolution}
\end{figure}

Due to the sensing uncertainty, we represent the map built by the robots as a 2D occupancy grid (OG) as opposed to a geometric map. An OG is a discretized representation of the environment where each cell represents the probability of that space being occupied. Figure~\ref{fig:finite_resolution}-Right shows a representative OG. Cells with a lower probability of occupancy ($<0.5$) are designated as free cells (represented as white in the OG) and cells with a higher probability of occupancy ($>0.5$) are designated as occupied cells (represented as black in the OG). Cells which have not been observed are marked as unknown (represented as gray in the OG).

Typical sensors such as cameras and laser rangefinders have a finite angular resolution. Consider three rays from the laser as shown in Figure \ref{fig:finite_resolution}-Left. The rays intersect obstacles at the cells marked as black and all the cells the ray intersect between the robot (marked in blue) and the cell are marked as free. Due to the finite resolution of the laser ($0.395^{\circ}$ for the Hokuyo laser used in our system), the gray cells, even though they are in the field of the laser, are unobserved and this leads to gaps in observations. This leads to false frontiers being detected and hence such erroneous frontiers have to be discarded. We employ a simple heuristic of checking the size of a candidate frontier and discard those below a threshold.

Consider the robot (and the laser) located at the blue circle in Figure~\ref{fig:finite_resolution}-Right. The red ray represents one of the laser rays. In order to check for blocking vertices, occupied cells with a neighboring frontier cell are shortlisted first. In the figure, the cells marked with yellow and red \emph{X} are identified. A blocking vertex, as defined earlier, is a reflex vertex. We can detect a reflex vertex in an OG by checking its four neighbors. If the four neighbors are an occupied cell, an unknown cell, a free cell which is a frontier, and a free cell which is not a frontier we mark it as a blocking vertex. In Figure~\ref{fig:finite_resolution}-Right, the cell marked with the yellow \emph{X} is detected as a blocking vertex.

\subsection{Information-Theoretic Subroutine} \label{sec:orient}
In the analysis for Algorithm 1, we assumed that we follow the shortest paths between A and B when executing the \texttt{goto(A,B)} subroutine. Instead, we can add local detours that will increase the information gain in order to improve the quality of the map. In order to maintain a constant competitive ratio, the robot is given a certain budget. This is known as the orienteering problem~\cite{blum2007approximation}. The orienteering problem seeks to find a path of length no more than a given budget, that maximizes the reward collected along the path. The reward is a function of the set of vertices visited along the path. If the reward is modular, then there exists a constant-factor approximation for the problem~\cite{blum2007approximation}.

Our objective is to minimize the uncertainty in the map. Mutual information~\cite{charrow2015information} is a suitable reward function that can be used. Mutual information measures the expected reduction in the entropy (i.e., uncertainty) of the map based on the measurements it expects to receive along the path. Mutual information is known to be submodular and monotonically increasing~\cite{krause2008near}.\footnote{Strictly speaking, mutual information is only monotonically increasing when the number of vertices on the path is much less than the number of vertices on the graph~\cite{krause2008near}.} We use the recursive greedy algorithm for submodular orienteering by Chekuri and Pal~\cite{chekuri2005recursive} to add local detours to the path. Chekuri and Pal~\cite{chekuri2005recursive} give a quasi-polynomial time recursive greedy algorithm that yields an $\mathcal{O}(\log \text{OPT})$ approximation for this problem, where $\text{OPT}$ is the optimal reward.

\begin{figure*}[thb]
\centering
\includegraphics[width=0.2\textwidth]{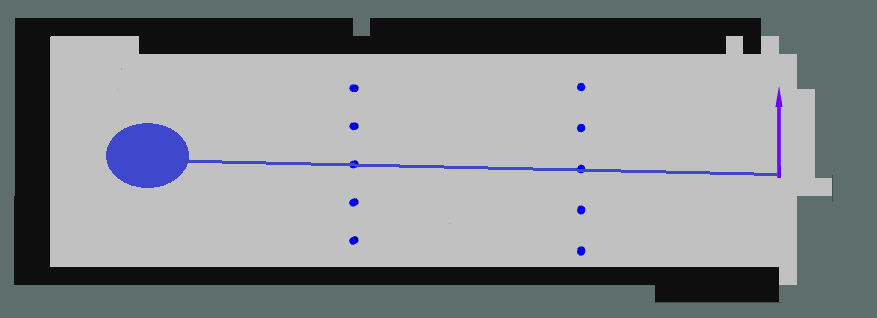}
\includegraphics[width=0.2\textwidth]{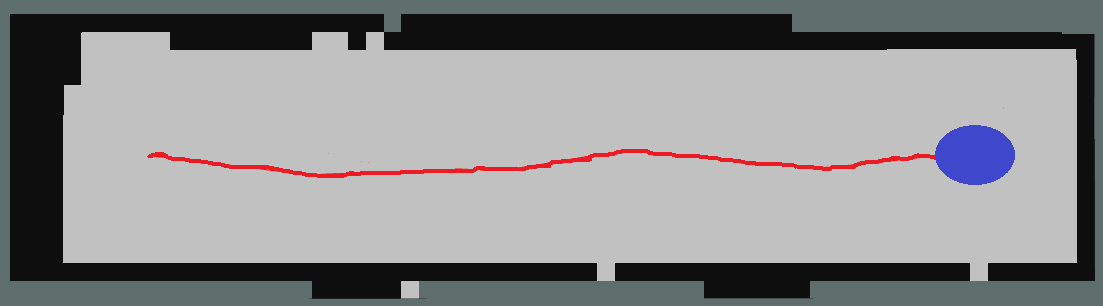}
\includegraphics[width=0.2\textwidth]{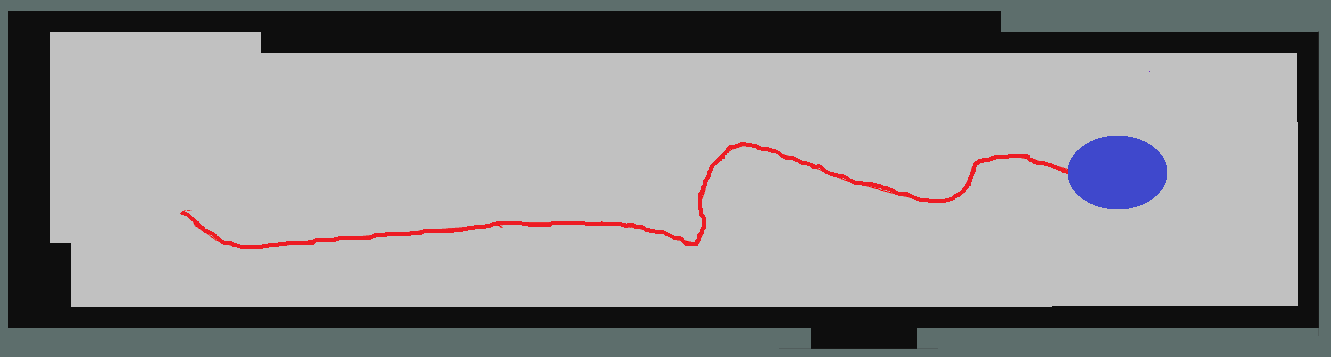}
\includegraphics[width=0.2\textwidth]{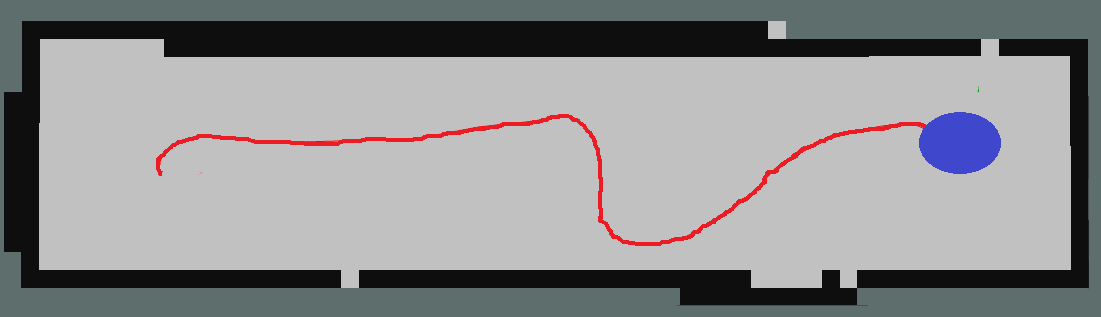}
\caption{The first figure shows the input graph imposed on the shortest path (blue line) between $s$ and $t$ for a small, test case. The other figures show the variation of path taken with change in budget ($d$, $2d$, and $4d$, where $d$ is the length of the shortest path).\label{fig:orient}}
\end{figure*}

The orienteering subroutine takes as input: an input graph, $G$, a start vertex, $s$, a goal vertex, $t$, a budget, $B$, a set of nodes visited, $X$ and returns a path that maximizes a submodular reward function subject to the budget. The input graph is generated by imposing a grid (add points above and below the path) on the shortest path between $s$ and $t$ as shown in Figure~\ref{fig:orient}. The start vertex is the current position of the robot, and the goal vertex is the next node in the tree to be visited. We set the budget to $B=\alpha d$, where $d$ is the length of the shortest path from $s$ to $t$, and $\alpha\geq 1$ is any constant. As $\alpha$ increases, we expect the quality of the map to improve at the expense of the cost of exploration (Figure~\ref{fig:orient}).

\begin{theorem}
\label{Orienteering}
Let $B=\alpha d$ be the assigned budget to the algorithm, i.e., the robot is allowed to take a path of cost $\alpha\geq 1$ times the cost of the shortest path, $d$, between the nodes. Let $C_\text{EXPLORE-INFO}^{p}$ be the cost of exploring the the polygon using the proposed algorithm along with the information theoretic subroutine. We have,
\begin{align} \label{cost_orient}
\frac{C_\text{EXPLORE-INFO}^{p}}{C_\text{OPT}^p}\le \frac{2\alpha(\sqrt2{p} +{\log p})}{1+\log p}
\end{align}
\end{theorem}
\begin{proof}
From Theorem~\ref{Exploration}, we have:
\begin{align}
 \frac{C_\text{EXPLORE}^{p}}{C_\text{OPT}}&\le \frac{2(\sqrt2{p} +{\log p})}{1+\log p}
\end{align}
 Each $s-t$ path between nodes is allotted a budget $B$, hence the total cost of our algorithm $C_\text{EXPLORE}^{p}$ is multiplied by at most a factor of $\alpha$. Hence the total cost of the algorithm with the information theoretic subroutine is given by,

\begin{align} 
C_\text{EXPLORE-INFO}^{p} = \alpha C_\text{EXPLORE}^{p}
 \end{align}
Hence, we have,

\begin{align} 
\frac{C_\text{EXPLORE-INFO}^{p}}{C_\text{OPT}^{p}}\le \frac{2\alpha(\sqrt2{p} +{\log p})}{1+\log p}
 \end{align}
 
\end{proof}

We modify the subroutine in order to save computation time by restricting the paths to be strictly moving forward, i.e, the path cannot have edges moving back towards $s$. This reduces the computation time by an order of magnitude. This is a heuristic and therefore the approximation ratio may not necessarily hold for the heuristic.

\subsection{General Environments}
Our algorithm also works for environments which are not orthogonal when occupancy grids are used as the underlying representation. Occupancy grids, typically, are orthogonal polygons by construction. Furthermore, for environments with holes (i.e., obstacles) the algorithm would create a tree and explore this tree. While this is correct, the distance traveled by the robots can be much higher than the optimal cost and as such the competitive ratio does not hold.  

In the next section, we evaluate the empirical performance of our algorithm through simulations in such scenarios.

\section{Evaluation}
In this section, we report results from simulations and hardware experiments. 
\subsection{Simulations} \label{sec:sims}
\begin{figure*}[thb]
\centering
\includegraphics[height=1.4cm]{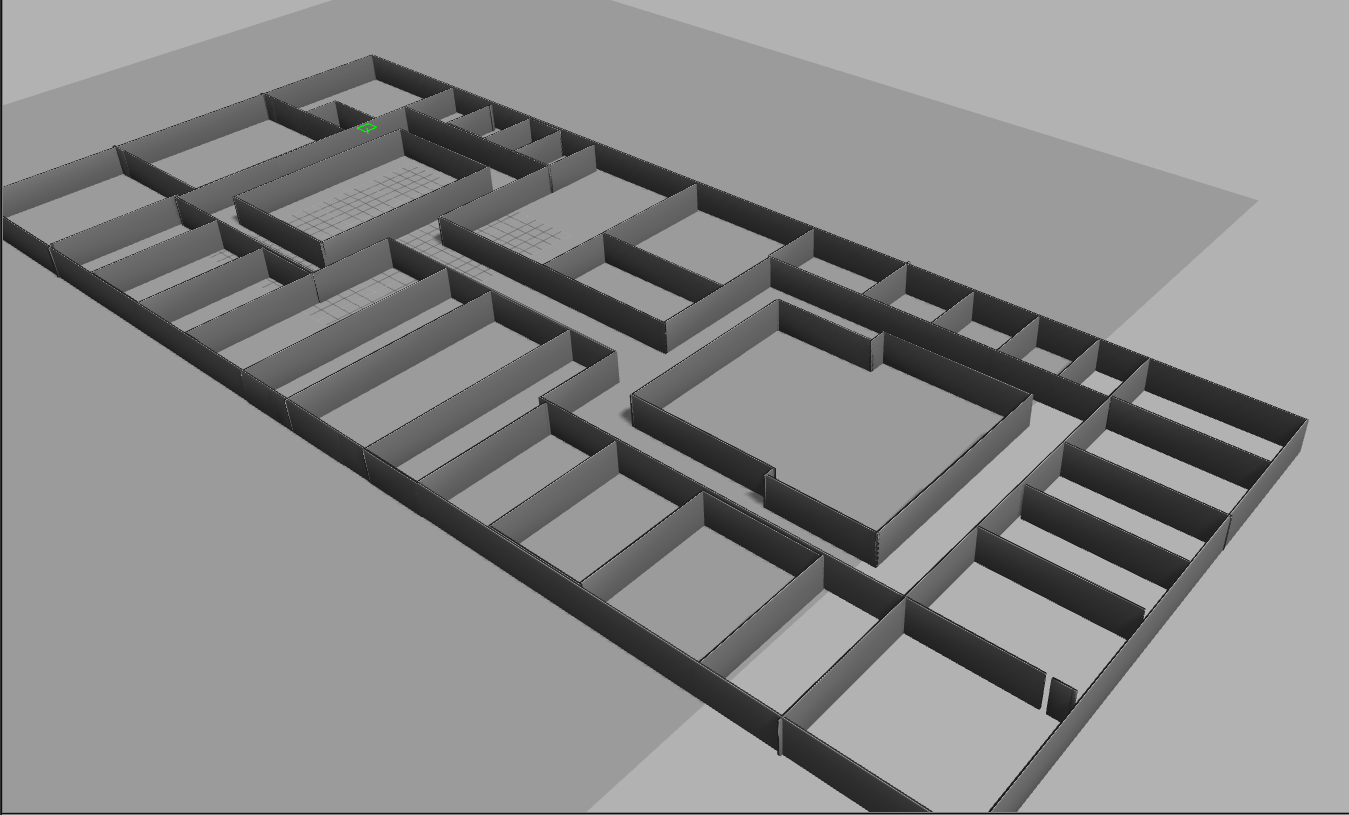}
\includegraphics[height=1.4cm]{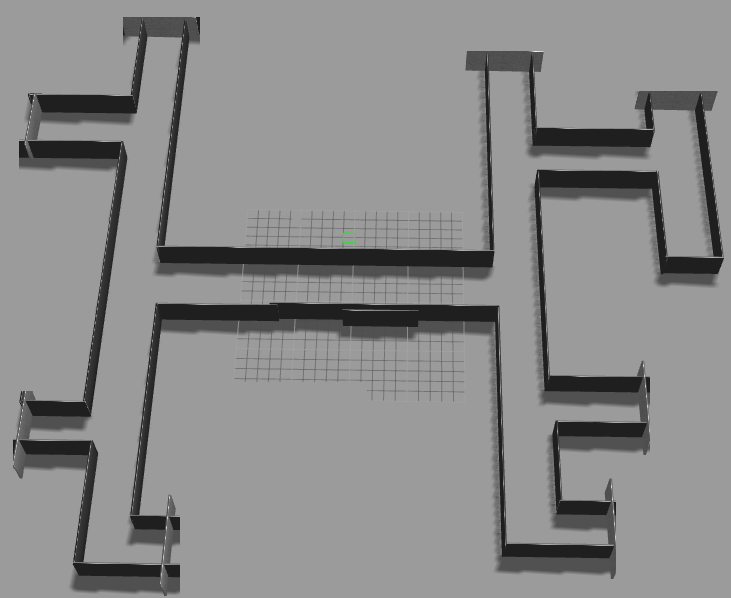}
\includegraphics[height=1.4cm]{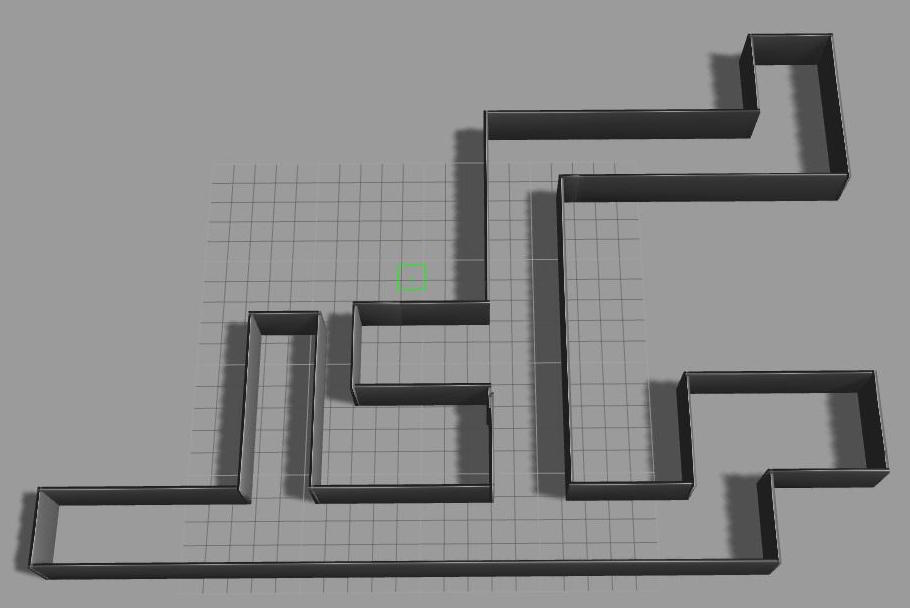}
\includegraphics[height=1.4cm]{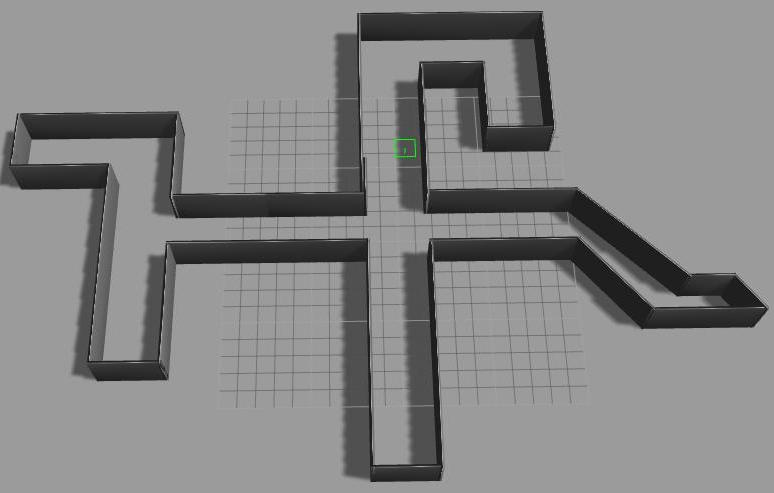}
\includegraphics[height=1.4cm]{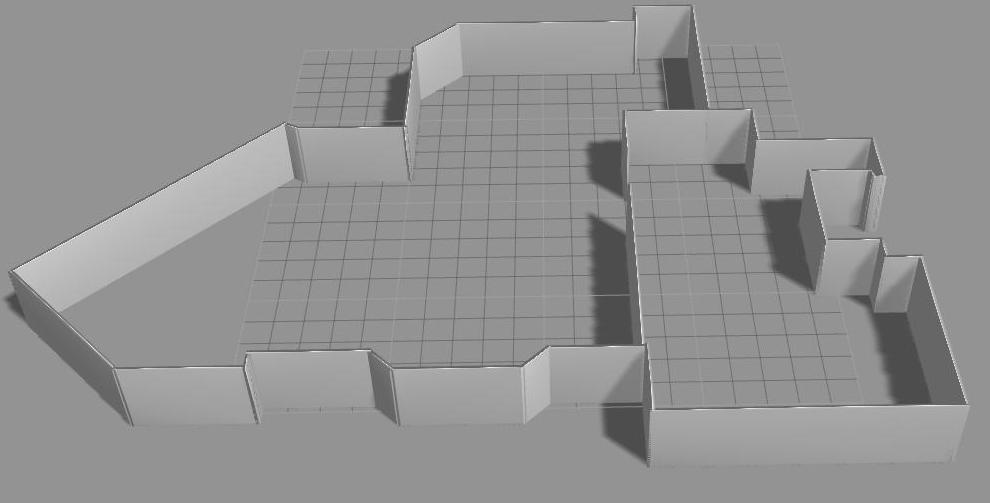}
\caption{Simulation environments 1 -- 5 from left to right.\label{fig:gazebo}}
\end{figure*}

We implemented our algorithm using ROS~\cite{quigley2009ros} and carried out simulations in Gazebo~\cite{koenig2004design} in order to verify the correctness of the exploration algorithm in realistic environments. The five Gazebo simulation environments used (Figure~\ref{fig:gazebo}) are not all orthogonal and simply-connected -- assumptions only required for the analysis to hold. The simulated robot is a differential-drive Pioneer P3-DX robot with a 2D Hokuyo laser range finder with a maximum range of 5 meters. The robot is localized in the environment using the \emph{amcl}~\cite{amcl} package using a pre-defined map. Mapping during exploration is done using the octomap ROS package. The laser scan is converted into a point cloud which is then fed as input to the \emph{octomap}~\cite{hornung2013octomap} node. Our implementation is available online at~\url{https://github.com/raaslab/Exploration}.

%

\begin{figure*}[thb]
\centering
\includegraphics[width=0.32\textwidth]{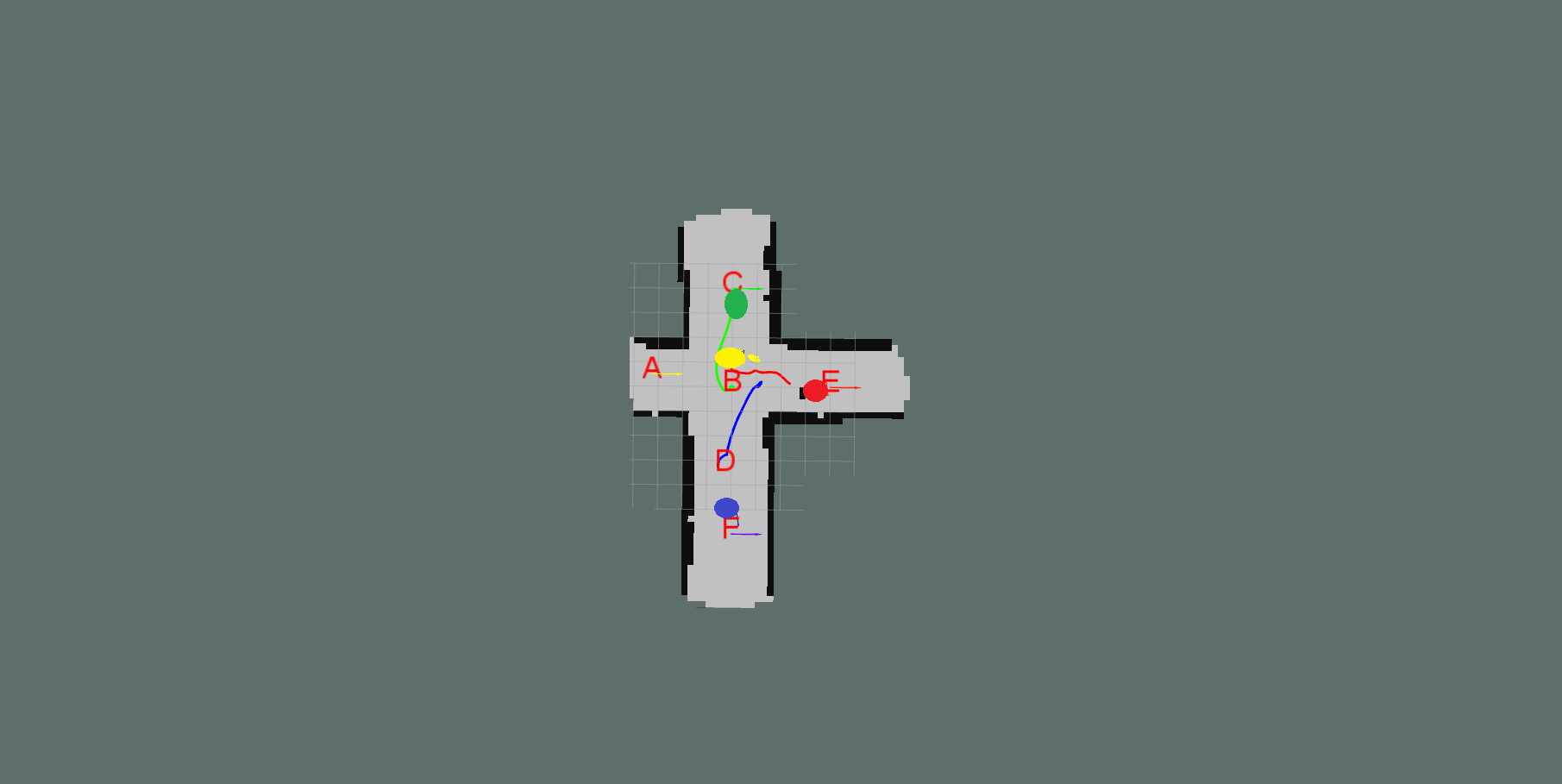}
\includegraphics[width=0.32\textwidth]{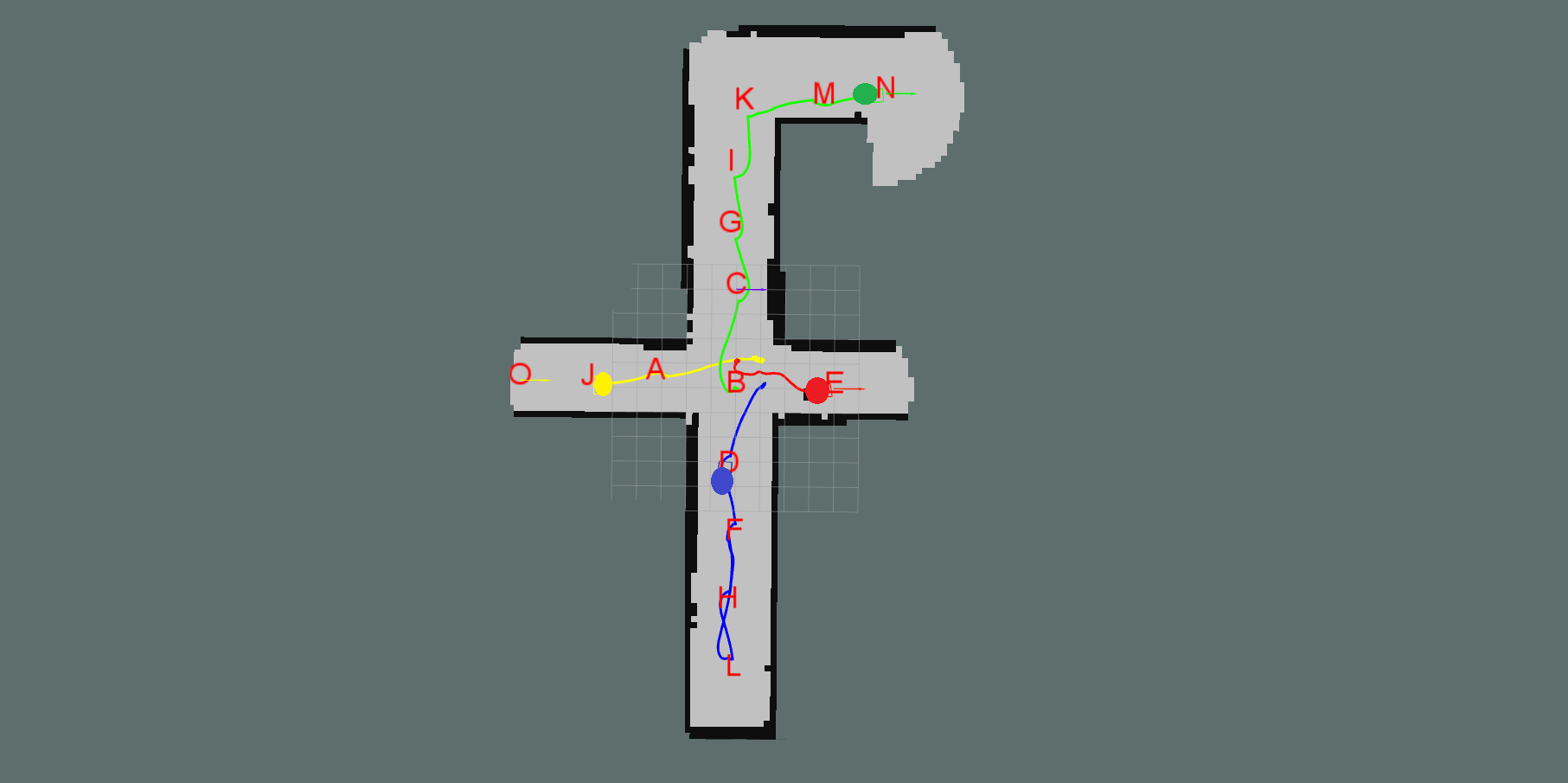}
\includegraphics[width=0.32\textwidth]{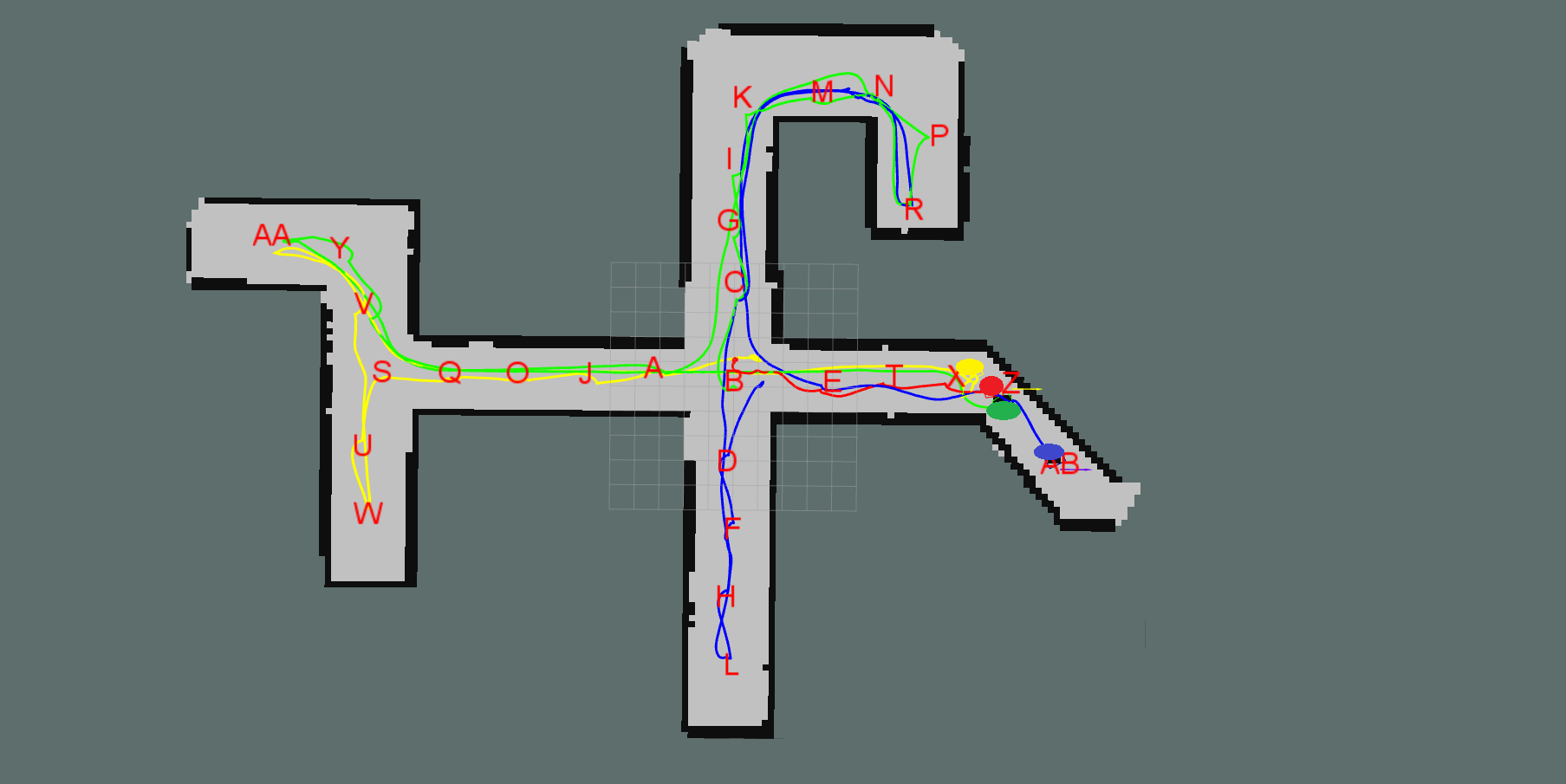}
\includegraphics[width=1.0\textwidth]{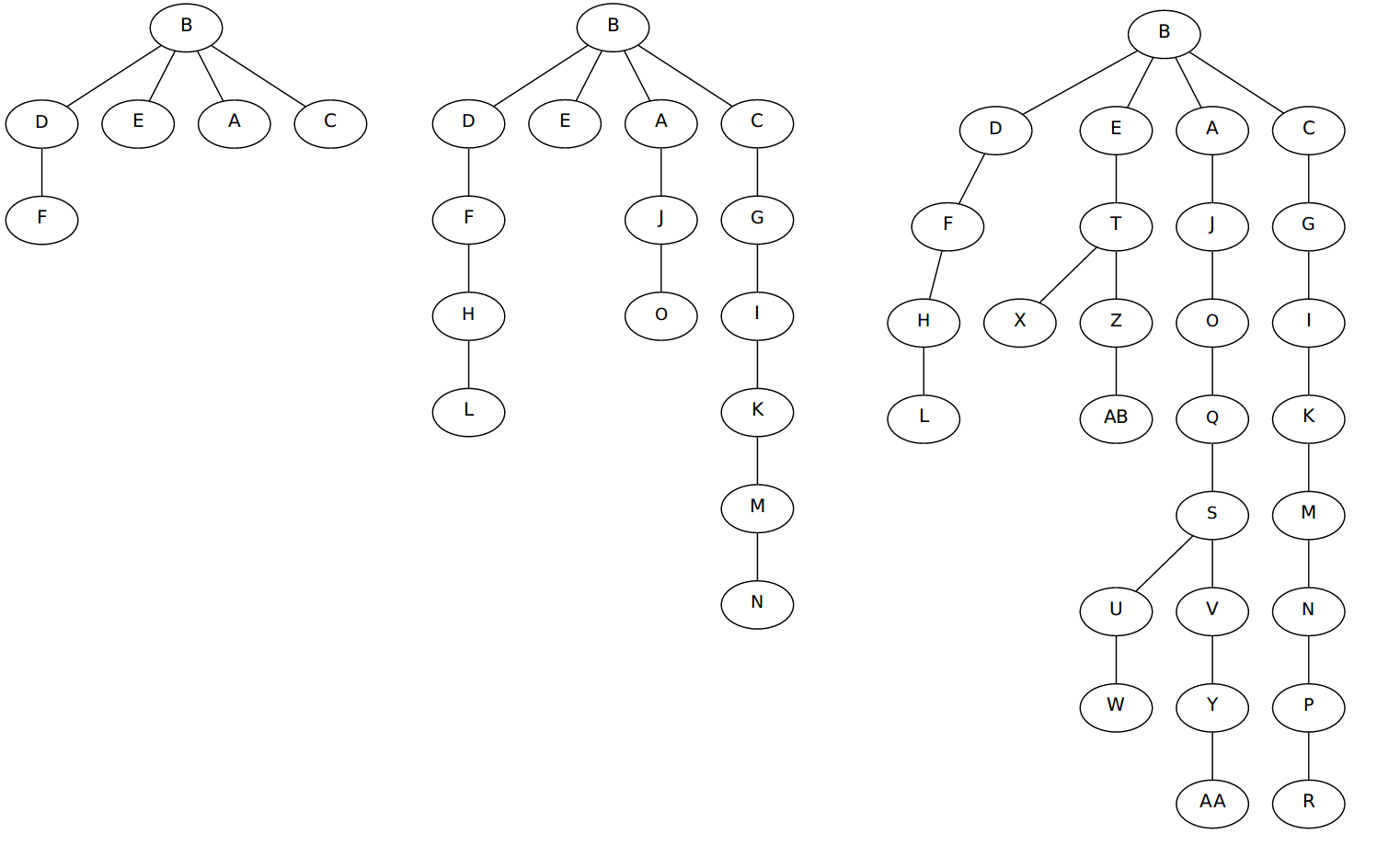}
\caption{Various stages while exploring environment 3 using four robots. The map explored and built along with the corresponding trees are shown.\label{fig:sims3}}
\end{figure*}

\begin{figure*}[htb]
\centering
\includegraphics[height=1.5in]{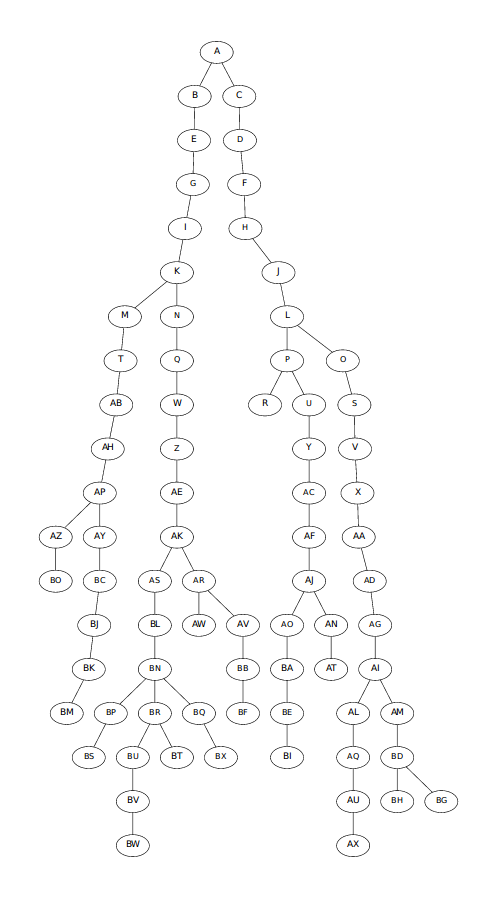}
\includegraphics[height=1.5in]{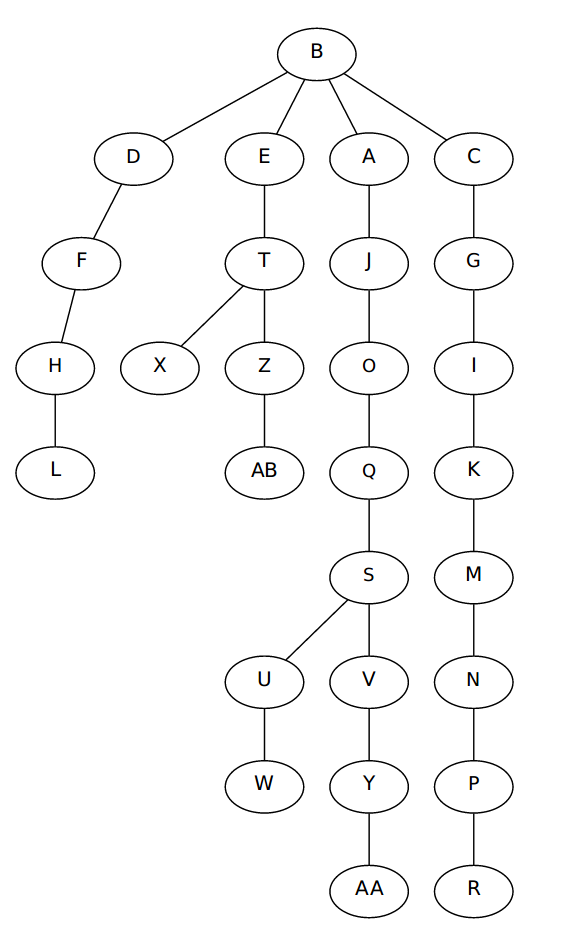}
\includegraphics[height=1.5in]{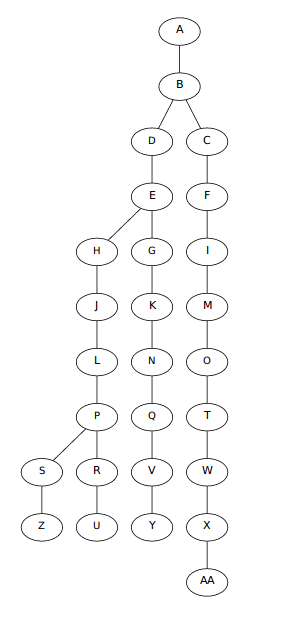}
\includegraphics[height=1.5in]{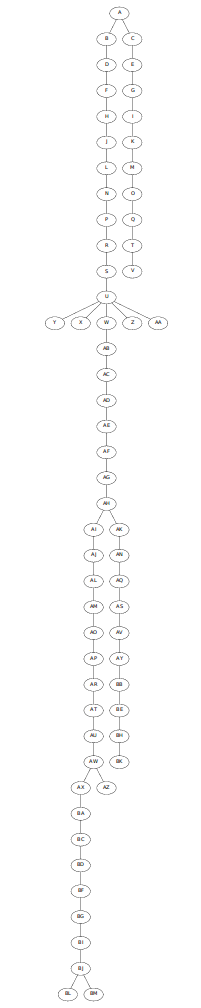}
\includegraphics[height=1.5in]{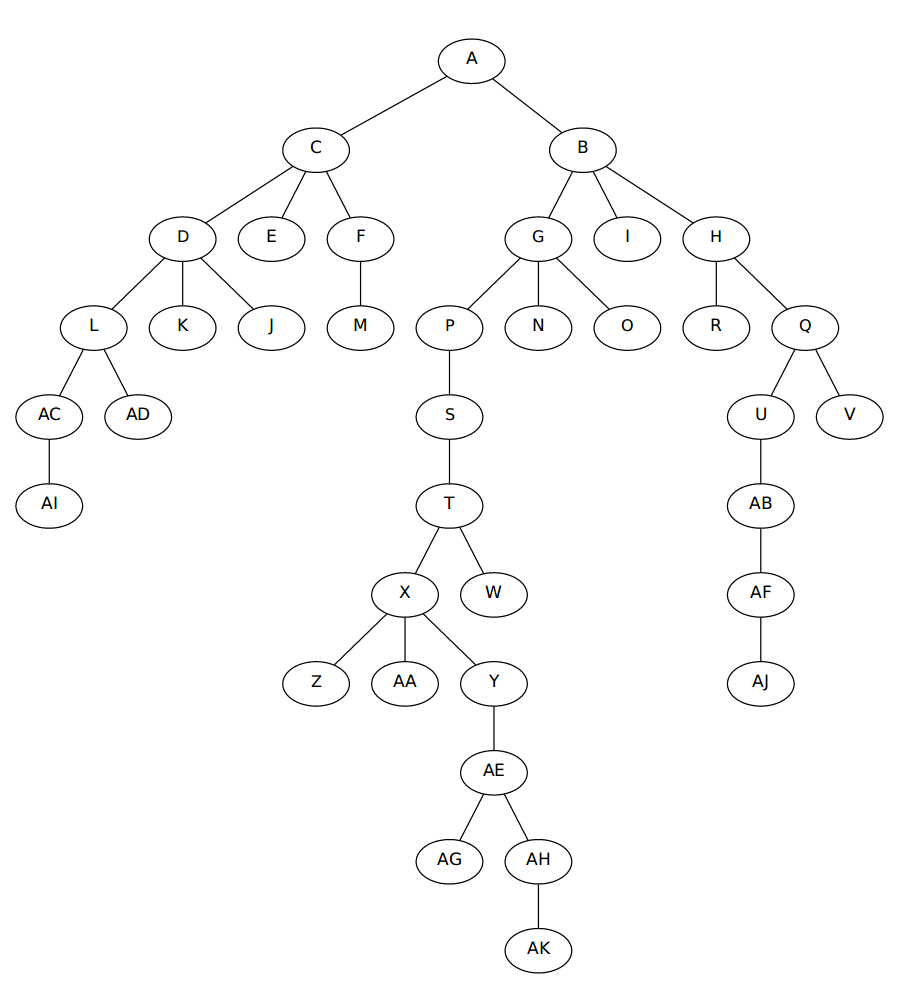}
\caption{Final tree produced after exploring the five environments in Figure~\ref{fig:gazebo} with four robots. The maximum distance traveled by a robot during exploration is given in Table~\ref{tab:results}\label{fig:trees}.}
\end{figure*}

%

Figure~\ref{fig:sims3} shows various stages of exploration with four robots. The figure also shows the partial exploration tree built by the algorithm. The final trees produced while exploring all the environments are given in Figure~\ref{fig:trees}. Table~\ref{tab:results} shows the maximum distance traveled by a robot (in meters) during exploration for all the environments.

\begin{table}
\begin{center}
    \begin{tabular}{ | l | c | c | c | c |c |}
    \hline
     & \textbf{Env. 1} & \textbf{Env. 2} & \textbf{Env. 3} & \textbf{Env. 4} & \textbf{Env. 5}\\ \hline
    \textbf{1 Robot} & 214.11 & 362.29 & 124.77 & 135.62 & 156.03\\ \hline
    \textbf{2 Robots} & 121.95 & 223.50 & 76.87 & 82.69 & 74.50\\ \hline
    \textbf{4 Robots} & 127.78 & 152.18 & 83.39 & 63.51 & 64.36\\ \hline
    \end{tabular}
\end{center}
\caption{Maximum distance traveled by a robot to explore environments in Fig.~\ref{fig:gazebo}.\label{tab:results}}
\end{table}

The cost of exploring environments 1, 3, 4, and 5 remains almost the same when the number of robots is increased from two to four. This is due to the fact that the exploration tree is not a balanced tree (Figure~\ref{fig:trees}). On the other hand, in environment 2, the tree contains four or more \emph{under-exploration} branches at all times. Consequently, the cost of exploration decreases significantly when four robots are used as opposed to just two.

The effect of the budget on the entropy is shown in table~\ref{tab:results2}. We can see that increase in budget reduces the overall entropy in the environment but at the cost of increased exploration time. We can also see that increasing the number of robots also decreases the entropy in the environment. 

\begin{table}
\begin{center}
    \begin{tabular}{ | l | c | c | c | c |}
    \hline
     & \textbf{Budget} & \textbf{Entropy}  & \textbf{Total Time}\\ \hline
    \textbf{1 Robot} & 1 & 24713 & 1131\\ \hline
    \textbf{1 Robot} & 2 & 24479 & 2719\\ \hline
    \textbf{1 Robot}& 4 & 24131 & 6187\\ \hline
    \textbf{2 Robots} & 1 & 24400 & 949\\ \hline
    \textbf{2 Robots} & 2 & 24267 & 3406\\ \hline
    \textbf{2 Robots} & 4 & 24039 & 6584\\ \hline
    \end{tabular}
\end{center}
\caption{Effect of budget while exploring first environment in Fig.~\ref{fig:gazebo}.\label{tab:results2}}
\end{table}

\subsection{Experiments} \label{sec:expt}

\begin{figure}[htb]
\centering
\includegraphics[height=1.5in]{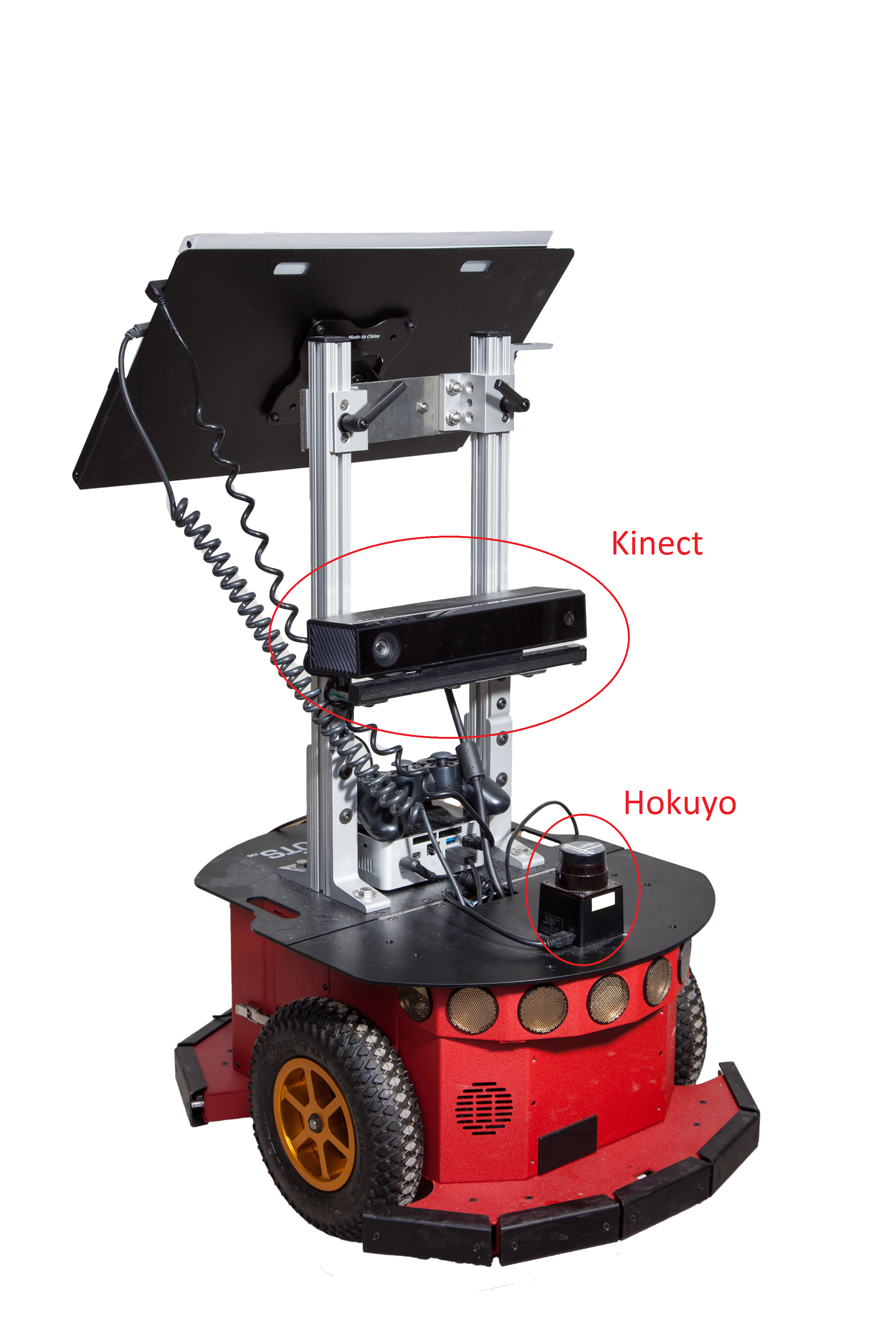}
\includegraphics[width=0.45\textwidth]{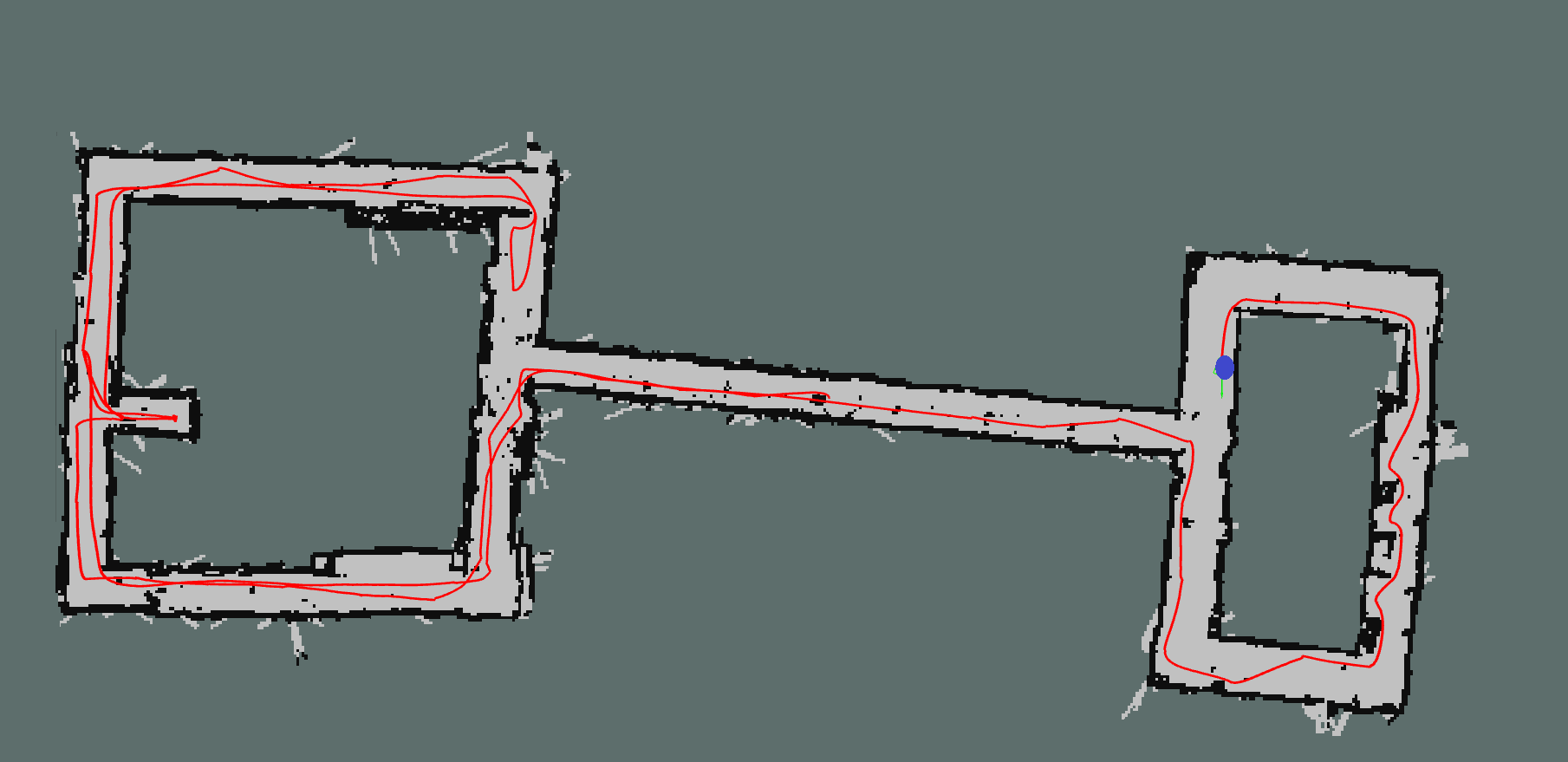}
\caption{\textbf{(Top)} Pioneer P3-DX with the Hokuyo laser used for the experiments. The 2D Hokuyo laser was used for robot localization and the Kinect2 sensor was used for mapping during exploration. \textbf{(Down)} Final map as well as the exploration path during the corridor experiment.\label{fig:robot}}
\end{figure}

We carried out experiments using a Pioneer P3-DX robot mounted with a Hokuyo URG-04LX-UG01 2D laser and a Kinect2 RGBD sensor (Figure~\ref{fig:robot}-Top). The laser was configured to use 180$^{\circ}$ field of view with a resolution of 0.395$^{\circ}$. During the exploration experiments, the robot used the 2D laser for localization on a pre-built map. The map was generated using RTAB-map~\cite{labbe2014online,labbe2013appearance} and the localization was carried out using the \emph{amcl} package from ROS. Note that having a pre-built map is not a requirement for our algorithm. The amcl localization component could be replaced by, for example, any SLAM implementation. Furthermore, the robots generated a new map during the exploration process using octomapping~\cite{hornung2013octomap} with the Kinect2 sensor. This map (and not the pre-built map) was used as the basis for finding blocking vertices in the proposed exploration algorithm.

\begin{table}
\begin{center}
    \begin{tabular}{ | l | c | c | c | c |}
    \hline
     & \textbf{Budget} & \textbf{Entropy} &   \textbf{Total Time(s)}\\ \hline
    \textbf{1 Robot} & 1 & 78493 & 1516 \\ \hline
    \textbf{1 Robot} & 2 & 78128 & 2453 \\ \hline
    \textbf{1 Robot}& 4 & 77917 & 5427 \\ \hline
    \end{tabular}
\end{center}
\caption{Effect of budget while exploring the corridor shown in Fig. \ref{fig:robot} using the real robot \label{tab:results3}}
\end{table}
\begin{figure}[htb]
\centering
\includegraphics[width=0.45\textwidth]{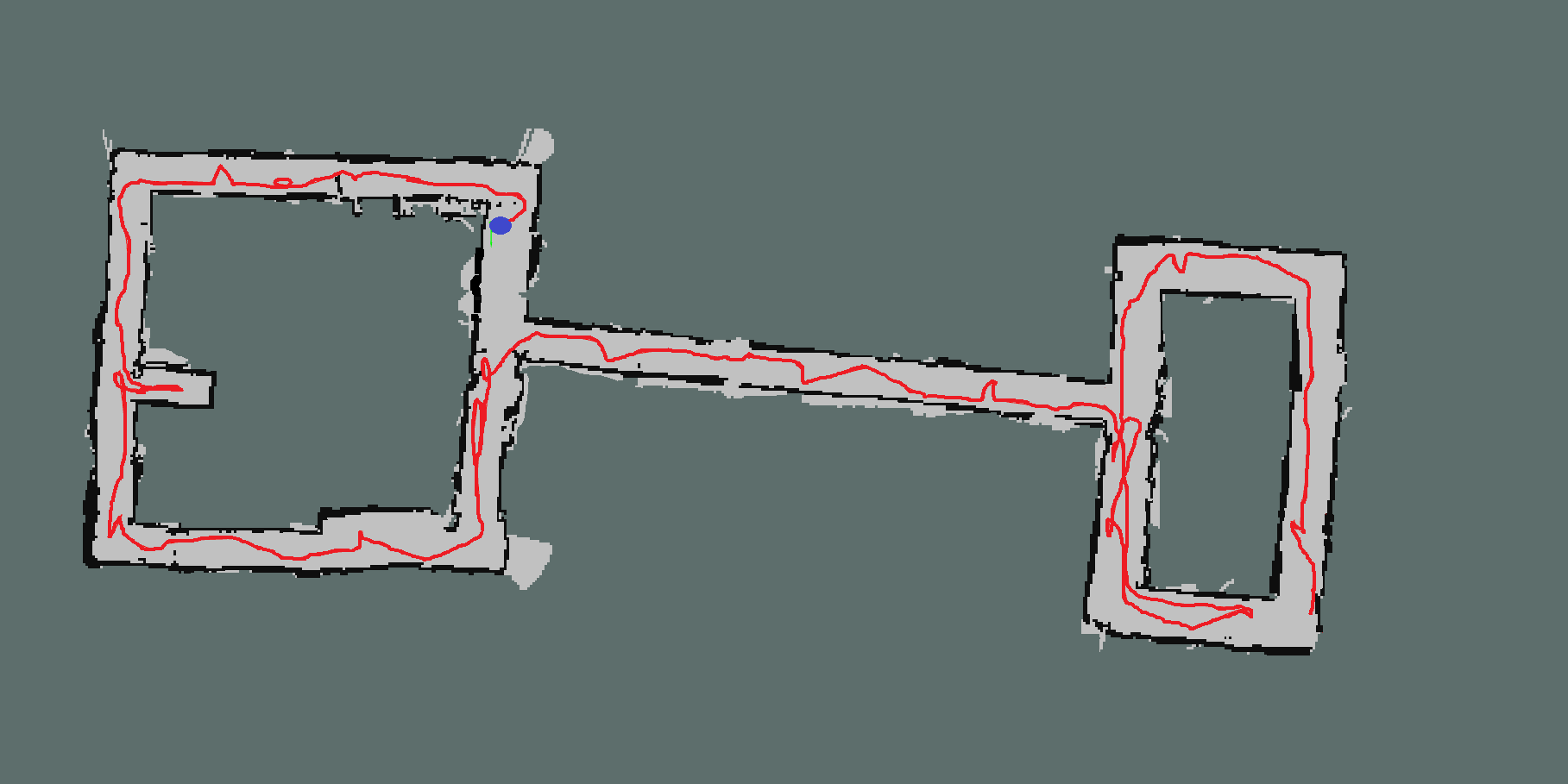}
\caption{Map of sixth floor of Whittemore hall built and the path taken by the robot after exploration of the environment with a budget of 2.}
\label{fig:robot_expt_orient}
\end{figure}



The mapping and localization were performed on the pioneer P3-DX robot with two onboard computers in a master/slave configuration. The slave i7 Intel NUC was dedicated to processing the Kinect2 data and the master i7Intel NUC was dedicated to running the localization and exploration algorithm. Figure~\ref{fig:robot} shows the results of an exploration experiment in a corridor environment along with the path followed by the robot. This proof-of-concept experiment shows that the algorithm can be applied in real world scenarios. A video of the system in operation is available online at \url{https://github.com/raaslab/Exploration}
\section{Conclusion} \label{sec:conc}
We presented an algorithm for exploring an unknown polygonal environment using a team of $p$ robots in the least amount of time. Our main theoretical contribution was to show that if the underlying environment is an orthogonal polygon without holes then our algorithm yields a constant competitive ratio for fixed $p$. Next, we showed how to adapt our algorithm so that it can extend to real-world sensing constraints. We then leveraged the mutual information at different locations in the map to reduce the overall entropy in the map whilst maintaining a constant competitive ratio. We verified the behavior of our algorithm through simulations and experiments.

Future work includes extending our analysis to more general environments. Handling general polygons without holes, not necessarily orthogonal, is a direct extension of the algorithm presented here. The notion of blocking vertices remains the same and the underlying graph will still be a tree. However, the \emph{extension goal} corresponding to the blocking vertex needs to be carefully defined. An immediate avenue of future work is to leverage the algorithm from~\cite{deng1998learn} that allows for obstacles in orthogonal environments. For polygons with holes, the underlying graph is no longer a tree. Hence, a general graph exploration algorithm would have to be used. 

\bibliographystyle{plain}
\bibliography{refs}
\end{document}